\documentclass[11pt]{article}
\usepackage{acl}

\usepackage{amsmath,amssymb,mathtools}
\usepackage{booktabs}
\usepackage{graphicx}
\usepackage{multirow}
\usepackage{array}
\usepackage{enumitem}
\usepackage{makecell}
\usepackage{tabularx}
\usepackage{colortbl}
\usepackage{microtype}
\usepackage{xspace}
\usepackage{url}
\usepackage{adjustbox}
\usepackage{float}
\usepackage{listings}
\usepackage{algorithm}
\usepackage{algpseudocode}

\graphicspath{{figures/}}
\setlist[itemize]{leftmargin=*,topsep=2pt,itemsep=1pt,parsep=1pt}
\setlist[enumerate]{leftmargin=*,topsep=2pt,itemsep=1pt,parsep=1pt}
\newcommand{\medisim}{\textsc{Medi-Sim}\xspace}
\newcommand{\alphaevolve}{\textsc{AlphaEvolve}\xspace}
\newcommand{\openevolve}{\textsc{OpenEvolve}\xspace}
\newcommand{\drg}{\textsc{DRG}\xspace}
\newcommand{\kpi}{\textsc{KPI}\xspace}
\newcommand{\cmi}{\textsc{CMI}\xspace}

\definecolor{l3alpha}{RGB}{238,247,255}
\definecolor{l3best}{RGB}{197,232,207}
\definecolor{l3risk}{RGB}{255,213,181}
\newcommand{\lthreemethod}[1]{\cellcolor{l3alpha}#1}
\newcommand{\lthreebestcell}[1]{\begingroup\setlength{\fboxsep}{1pt}\colorbox{l3best}{\textbf{#1}}\endgroup}
\newcommand{\lthreeriskcell}[1]{\begingroup\setlength{\fboxsep}{1pt}\colorbox{l3risk}{#1}\endgroup}
\newcommand{\tabrisk}[1]{\lthreeriskcell{#1}}
\newcommand{\tabgood}[1]{\lthreebestcell{#1}}

\title{Healthcare Mechanisms from Policy-as-Code Search under Strategic Provider Response}
\author{
Zihan Wang\textsuperscript{1} \quad
Xiang Xu\textsuperscript{1} \quad
Hongyuan Zha\textsuperscript{1} \quad
Wenhao Li\textsuperscript{2} \\[0.3ex]
\textsuperscript{1}The Chinese University of Hong Kong, Shenzhen \\
\textsuperscript{2}Tongji University
}

\begin{document}
\maketitle
\raggedbottom

\begin{abstract}
Healthcare mechanisms are inseparable from the strategic provider response they induce: existing healthcare AI benchmarks hold this response fixed and so cannot evaluate mechanisms by the equilibrium they produce. 
We recast hospital mechanism design as program synthesis for language models: typed, inspectable rule programs are executed and scored by \medisim{}, a multi-agent simulator with five strategic provider channels (coding, selection, delay, effort, triage). 
An incentive sweep recovers classical health-economics findings as adjacent regimes---up-coding and low-complexity-patient selection under profit pressure, and Goodhart-style drift where measured performance becomes anti-correlated with true outcomes---and a single audit lever exposes \emph{pressure migration}: closing the coding channel more than doubles low-complexity selection. 
LLM-guided evolutionary code search over the same rule-program space then synthesizes an inspectable mixed-objective program that eliminates up-coding, halves rejection, and retains most of the profit-oriented baseline's funds.
\end{abstract}

\section{Introduction}\label{sec:intro}

A case-based payment becomes a coding rule, an audit reshapes patient selection, and a quality bonus redirects effort toward the measured score: in each case, a hospital mechanism is realized as the \emph{composition} of an administrator's instruction with a provider's best response, and the composition---not the text---determines billing, access, and outcome.\footnote{Healthcare-specific terminology used throughout the paper---including hospital \drg and \drg-style arrivals, hospital \cmi, hospital \kpi and \kpi steering, the five provider-response channels (coding, selection, delay, effort, triage), the coding and measurement wedges, gold-plating, skimping, cream-skimming, the Identify--Produce--Settle (IPS) loop, and the hospital policy DSL---is collected in the glossary of Appendix~\ref{sec:app-terms-arrivals}.} 
The dynamic we focus on is \emph{pressure migration}---a multi-channel feature of strategic best response in which, when a rule closes one provider channel, the same incentive resurfaces in an adjacent one, so a benchmark that scores rules against a fixed provider systematically over-rewards mechanisms whose effect is to relocate rather than remove distortion. 
We therefore evaluate hospital mechanisms inside a closed-loop strategic-response simulator, and because regulated deployment additionally requires every rule to remain auditable line by line, we restrict the administrator's policy class to inspectable, typed rule programs---reframing mechanism design as \emph{program synthesis} over a constrained administrative interface.

Pressure migration is visible across three decades of healthcare reform. Hospitals respond to Medicare diagnosis-pricing changes by re-coding rather than by treating more patients \citep{dafny2005hospitals}; Medicare Advantage risk scores grow faster than fee-for-service scores through coding intensity \citep{kronick2014measuring}; and English NHS waiting-time targets change both reported waits and the operations that produce them \citep{bevan2006measured,propper2010incentives}. The machine-learning reading is direct: each is \emph{Goodhart-style drift} \citep{manheim2018goodhart} delivered through a strategic-response shift in the data-generating process, of the kind formalized by strategic classification and performative prediction \citep{hardt2016strategic,perdomo2020performative}.

Existing benchmarks cannot see this dynamic. Healthcare AI environments train clinician-level policies on a fixed environment with passive providers \citep{komorowski2018ai,yu2021reinforcement,gottesman2019guidelines}, treating provider behavior as exogenous noise. Automated mechanism-design systems do model strategic response, but instantiate taxes, auctions, or generic allocation rather than healthcare primitives---reimbursement, audits, care-team queues, measured quality---and their searched controllers are black-box neural networks that fail line-by-line auditability \citep{zheng2022ai,dutting2024mechanism,sandholm2003automated}. Neither camp scores administrator rules and strategic provider responses through realized access, reimbursement, and performance in the same rollout.

We instantiate the missing loop in \medisim{}. Administrator rules are written as \emph{policy-as-code}: typed, executable expressions over a fixed set of approved levers (incentive coefficients, audit intensity, bonus pool, performance-score weights) that are auditable line by line \citep{rudin2019stop}, and expose the kind of clear, context-relevant information emphasized in good machine learning practice for medical devices (GMLP)\citep{fda2021gmlp}. Providers respond through five named channels---coding, selection, delay, effort, and triage---drawn from health economics \citep{ellis1998creaming,ma1994health,kuhn2008upcoding,holmstrom1991multitask}, and an Identify--Produce--Settle (IPS) loop keeps rules, responses, and outcomes in the same rollout, treating settlement (reimbursement, scores, bonuses) as part of the mechanism rather than a reporting layer. The same loop is the search interface: because candidates are typed code expressions over a small state-feature set, useful mutations are \emph{semantic} edits over code rather than gradient steps or random rewrites---a regime in which LLM-guided code search outperforms unguided genetic operators \citep{romera2024funsearch,lehman2022elm,novikov2025alphaevolve}, while line-by-line auditability rules out neural controllers. The language model therefore acts as a code-editing operator over the rule program under a safety-penalized closed-loop fitness; provider agents are parameterized response classes, not LLMs.

Three experiments close the loop. An incentive sweep recovers classical findings as adjacent regimes of one phase diagram---up-coding and low-complexity selection under profit pressure, balanced-interior Goodhart drift---and administrative lever sweeps expose pressure migration: audits shift pressure from coding to selection, while bonus pools and KPI-steered flex capacity reveal proxy and waiting-time failures. LLM-guided code search over the same rule interface refines a diverse warm-start library into an inspectable mixed-objective program that eliminates up-coding, halves rejection, and retains most of the profit-oriented baseline's funds; ablations show warm-start priors and LLM-guided refinement to be jointly necessary.

\paragraph{Contributions.}
\textbf{(1) An LLM-program-synthesis testbed for high-stakes mechanism design.} We recast provider-side mechanism design as LLM-guided program synthesis over a typed administrative DSL, in which neural controllers are excluded by audit requirements and the LLM acts as a code-editing operator on inspectable rule programs under safety-penalized multi-agent rollouts.
\textbf{(2) A closed-loop strategic-response benchmark.} We release \medisim{}, an Identify--Produce--Settle simulator that keeps administrator rules, five strategic provider channels, and realized access/reimbursement/performance in the same rollout, exposing the channel-level diagnostics needed to detect strategic-response distortion.
\textbf{(3) Pressure migration as a benchmark phenomenon addressable by LLM-guided code search.} Classical healthcare failures occupy adjacent regimes of one mechanism space, and LLM-guided refinement of a diverse warm-start library can reduce targeted manipulation while monitoring whether pressure reappears on adjacent channels; in the main held-out mixed-policy comparison, the searched program closes the coding channel without increasing rejection. Ablations attribute the effect jointly to priors and LLM code editing.

\section{Problem Formulation}\label{sec:formulation}
We model the hospital as a finite-horizon stochastic Stackelberg game over $T$ periods. The \emph{hospital administrator} is the leader, committing to a mechanism action $u_t$ at each step; the \emph{provider population} is the follower, drawn from a tractable response class $\Pi_P$ described below. Throughout, $J$ indexes care teams.

\paragraph{State and leader action.}
The hospital state $X_t$ collects funds $F_t$, congestion $Q_t$, per-team queues $\{\mathcal Q_{j,t}\}$, the previous-period \kpi vector, and reputation $\mathrm{Rep}_t$ (Eq.~\eqref{eq:hospital-state}, App.~\ref{sec:app-formulation}). The leader action $u_t$ collects incentive coefficients $(\alpha_t,\beta_t)$ for provider financial and quality sensitivity, total and flexible capacities $(B^{\mathrm{tot}}_t,B^{\mathrm{flex}}_t)$, the bonus pool $B^{\mathrm{pool}}_t$ and softmax sharpness $\kappa$, the \kpi weights $(w_H,w_W,w_{\mathrm{rej}},w_C)$ on health/waiting/rejection/cost, audit intensity $q_t$, and an optional \kpi-steering switch $\xi_t$ (Eq.~\eqref{eq:leader-action}, App.~\ref{sec:app-formulation}).

\paragraph{Follower action: five distortion channels.}
Each team $j$ observes its queue, capacity signals, fatigue, and incentives, and chooses an action that decomposes into five channels:
\begin{equation}\label{eq:five-channels}
a_{j,t}=\big(\underbrace{\hat g_{i,t}}_{\textit{coding}},\ \underbrace{d^{\mathrm{acc}}_{ij,t}}_{\textit{selection}},\ \underbrace{d^{\mathrm{def}}_{ij,t}}_{\textit{delay}},\ \underbrace{E_{ij,t}}_{\textit{effort}},\ \underbrace{R_{ij,t}}_{\textit{triage/resource}}\big),
\end{equation}
indexed by candidate patient $i$. These are exactly the five channels through which providers strategically respond to medical mechanisms in the health-economics literature \citep{ellis1998creaming,ma1994health,dafny2005hospitals,kuhn2008upcoding,holmstrom1991multitask}, and they map one-to-one to the distortion measurements reported in \S\ref{sec:experiments}. The five-channel choice covers the main margins exposed by the Identify--Produce--Settle loop without making the response model too broad to diagnose channel-by-channel behavior. The provider-side team utility that drives these five channels is
\begin{equation}\label{eq:provider-util}
\begin{aligned}
U_{j,t} \;=&\; \alpha_t(\mathrm{Rev}_{j,t}-C_{j,t}) + \beta_t H_{j,t} \\
&{}+ \theta\,\mathrm{Bonus}_{j,t} - \nu\,\big[\max(0,\mathrm{Load}_{j,t}-\bar E_j)\big]^2,
\end{aligned}
\end{equation}
where $\theta>0$ is the fixed weight on the realized bonus and $\nu>0$ is the convex fatigue penalty above per-team load capacity $\bar E_j$.

\paragraph{Bounded-rationality response class.}
We restrict the follower to a tractable response class $\Pi_P=\{\pi_P^{\phi}:\phi\in\Phi\}$ parameterized by interpretable behavioral coefficients $\phi$ that govern, per channel, how aggressively the team's action moves with the local gradient of Eq.~\eqref{eq:provider-util}; the functional forms are given in \S\ref{sec:env}. This is a deliberate design choice rather than an equilibrium claim: it preserves per-channel identifiability, keeps each channel inspectable, and matches the comparative-static predictions used to validate the simulator in \S\ref{sec:experiments}.

\paragraph{Stackelberg objective.}
A mechanism is evaluated only through the rollout distribution induced by the follower's response. The leader optimizes a chosen social objective $o\in\{\mathrm{welfare},\mathrm{profit},\mathrm{mixed}\}$ with per-objective discounted return on seed $s$,
\begin{equation}\label{eq:leader-return}
G^o(\pi_A,\pi_P^*;s) = \sum_{t=1}^T \gamma_d^{t-1}\, r_A^o(X_t,u_t),
\end{equation}
and solves
\begin{equation}\label{eq:stackelberg}
\begin{aligned}
\pi_A^{o,*} \in \arg\max_{\pi_A\in\Pi_A}\big\{&\mathbb E_{s,\pi_P^*}[G^o]\\
&{}-\lambda_{\mathrm{unsafe}}\mathbb E[V]-\lambda_{\mathrm{var}}\mathrm{Var}_s[G^o]\big\},
\end{aligned}
\end{equation}
where $\pi_P^*\in\Pi_P(\pi_A)$ is the bounded-rationality best response, $V$ aggregates safety/distortion diagnostics (unsafe waiting, high-complexity deferral, up-coding, rejection, insolvency), and the variance term is a seed-reliability regularizer (App.~\ref{sec:app-formulation}). A mechanism is successful only if the induced provider behavior remains acceptable on every diagnostic, on the average seed \emph{and} reliably across seeds. We defer the choice of policy class $\Pi_A$ to \S\ref{sec:mac}, which discharges auditability constraints by instantiating $\Pi_A$ as a typed inspectable program class and solves Eq.~\eqref{eq:stackelberg} by \alphaevolve-style code search.

\section{The \medisim Environment}\label{sec:env}
\begin{figure*}[t]
  \centering
  \includegraphics[width=0.85\textwidth]{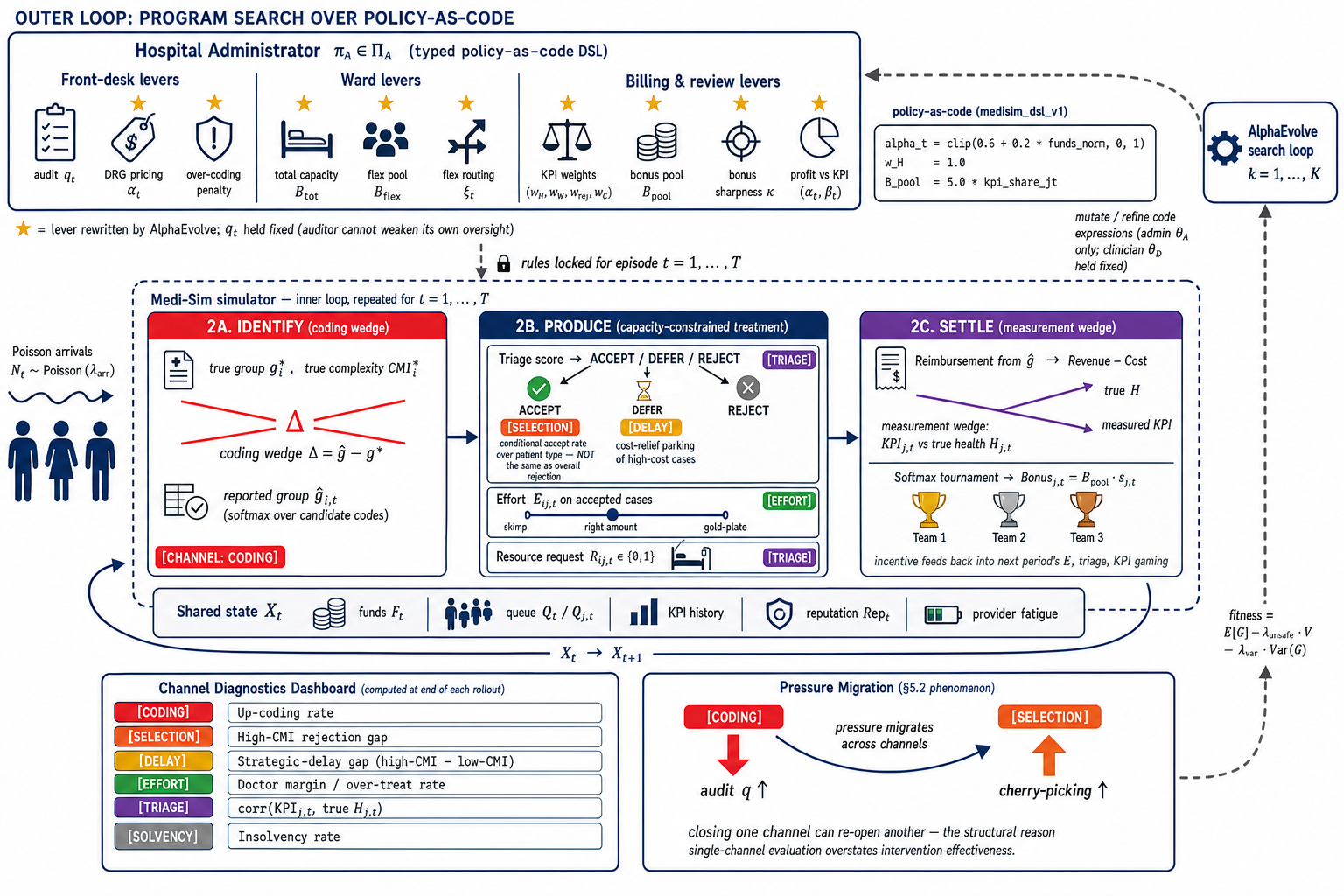}
  \caption{\medisim IPS and policy-as-code overview. Top: the hospital administrator writes episode-level front-desk, ward, and billing/review rules; stars mark levers refined by \alphaevolve. Middle: clinician programs respond within locked rules through the Identify--Produce--Settle loop. Bottom: the dashboard reports channel-level diagnostics that guide policy search.}
  \label{fig:overview}
\end{figure*}

\medisim{} implements the Identify--Produce--Settle (IPS) decomposition of \S\ref{sec:formulation} with the policy interface shown in Figure~\ref{fig:overview}: hospital administrative rules are fixed for an episode, clinician-side programs respond within those rules, and the resulting rollout dashboard exposes the same distortion channels used by L1--L3 (Algorithm~\ref{alg:sim-loop}, App.~\ref{sec:app-algorithms} gives the period-by-period loop). We describe each primitive in turn.

\subsection{Identify: arrivals, classification, and the coding wedge}\label{sec:identify}
At each step, a default hospital \drg-style arrival process\footnote{Hospital \drg-style arrivals carry clinical type, urgency, tolerance, and reimbursement-relevant case weight; see Appendix~\ref{sec:app-terms-arrivals}.} draws a Poisson batch of patients $\mathcal P_t$ with rate $\lambda_{\mathrm{arr}}$ (Eq.~\eqref{eq:arrival-poisson}, App.~\ref{sec:app-env}); non-Poisson kernels are admissible (App.~\ref{sec:app-terms-arrivals}). Each patient $i$ carries a true group $g_i^\star\in\mathcal G$, a normalized hospital case-mix index \cmi\footnote{\cmi is the normalized hospital case-mix index used as patient complexity and as the payment-relevant weight that coding can distort; see Appendix~\ref{sec:app-terms-arrivals}.} value $\mathrm{CMI}_i^\star\in[0,1]$, urgency $\mathrm{Urg}_i\in[0,1]$, and waiting tolerance. The default simulator uses an aggregated macro-\drg distribution inspired by common inpatient categories and \drg relative weights \citep{cms2026drg,hcup2024nis}. True complexity determines clinical need; coded complexity determines settlement. This split is the \emph{hospital coding wedge}\footnote{The hospital coding wedge is the gap between true clinical complexity and the reported billing group used for settlement; see Appendix~\ref{sec:app-terms-arrivals}.}: it creates the rent that the coding channel arbitrates.

\paragraph{Coding action.}
A hospital coder chooses a reported group $\hat g_{i,t}$ from a finite candidate set around the true group by applying a score-based choice rule to candidate groups. For each candidate group, the host simulator computes candidate-level signals: incremental reimbursement $\Delta R_i(g)$, audit-expected penalty under the configured audit schedule, ethics pressure from coded-complexity inflation, and the resulting coding gap. The coding rule maps these signals to a scalar candidate score; the baseline closed form is Eq.~\eqref{eq:coding-score} in App.~\ref{sec:app-env}.

The same score can be instantiated either as a stochastic softmax choice or as its deterministic zero-temperature limit, $\hat g_{i,t}=\arg\max_g s_i(g)$. The AlphaEvolve L3 implementation uses the deterministic variant to reduce evaluation noise, while keeping the candidate set, score features, audit schedule, and settlement routine fixed. As incentives vary, this score-based channel produces intermediate aggregate up-coding rates across patients and rollouts rather than making coding an all-or-nothing administrative switch.

In L3, \texttt{candidate\_score} is only the exposed behavioral scoring map for this channel. Edits to it reweight the coder's sensitivity to fixed host-computed features such as \texttt{upcode\_pressure}, \texttt{audit\_penalty}, \texttt{ethics\_pressure}, and \texttt{coding\_gap}; candidate construction, audit, penalties, and clawback remain host-side settlement routines.

\paragraph{Patient routing.}
Patients register to service units rather than being reassigned by the hospital. In L1/L2 active strategic routing is disabled and \kpi-aware steering acts only on the flexible capacity pool (\S\ref{sec:settle}), so selection and delay arise predominantly through the provider triage channel (\S\ref{sec:produce}); the external-validity implications are discussed in Limitations.

\subsection{Produce: capacity-constrained treatment}\label{sec:produce}
A service unit observes its queue, capacity signals, fatigue, and incentives. Its action realizes four of the five distortion channels at once: \emph{triage/resource} (accept, reject, defer, request a constrained resource), \emph{selection} (the conditional distribution of acceptance over patient types), \emph{delay} (deferral as a cost-relief lever), and \emph{effort} (intensity per accepted case). All four channels are implemented as closed-form behavioral rules that operate on the local gradient of the team utility $U_{j,t}$ defined in Eq.~\eqref{eq:provider-util}; collectively, these rules are the concrete instantiation of the response class $\Pi_P$ used in \S\ref{sec:formulation}.

\paragraph{Treatment production.}
For an accepted patient $i$ treated by unit $j$, true health output follows a diminishing-return production in effort $E_{ij,t}\ge 0$, gated by a constrained resource $R_{ij,t}\in\{0,1\}$ and scaled by team skill and inverse \cmi; cost is convex in effort with exponent $\phi>1$ (Eqs.~\eqref{eq:treatment}--\eqref{eq:cost}, App.~\ref{sec:app-env}). Effort therefore arbitrates the intensive margin (gold-plating versus skimping) while triage and selection arbitrate the extensive margin (who is treated and when), exactly as in the two-margin view of provider response \citep{ellis1998creaming,ma1994health}.

\paragraph{Triage and delay.}
Triage gates accept/reject/defer through a scalar score combining urgency, waiting, predicted margin, capacity, fatigue, and \kpi-targeting signals; strategic delay is the cost-relief channel that parks high-cost cases under profit pressure, producing the L1 delay signatures of \S\ref{sec:l1}.

\subsection{Settle: Hospital KPI, bonuses, and the measurement wedge}\label{sec:settle}
Team-level measured performance combines health with operational and financial penalties:
\begin{equation}
\begin{aligned}
\mathrm{KPI}_{j,t}=&\ w_H\bar H_{j,t}-w_W\bar W_{j,t}\\
&-w_{\mathrm{rej}}\mathrm{Reject}_{j,t}-w_C\bar C_{j,t}.
\end{aligned}
\end{equation}
Bonuses are allocated through a softmax tournament over $\mathrm{KPI}_{j,t}$ with sharpness $\kappa$ and pool $B^{\mathrm{pool}}_t$, yielding a local marginal bonus pressure $B^{\mathrm{pool}}_t\kappa\,s_{j,t}(1-s_{j,t})$ that feeds back into effort, triage, and \kpi-targeting behavior (Eqs.~\eqref{eq:bonus-softmax}--\eqref{eq:bonus-pressure}, App.~\ref{sec:app-env}). Because the measured hospital \kpi is a weighted aggregate of true health, waiting, rejection, and cost, it is generically misaligned with the principal's objective whenever complex care raises true health but worsens the proxy score---this is the \emph{hospital measurement wedge}\footnote{The hospital measurement wedge is the gap between true clinical value and the measured hospital \kpi used for bonuses or steering; see Appendix~\ref{sec:app-terms-arrivals}.} that hosts Goodhart-style gaming on triage signals \citep{holmstrom1991multitask,baker1992incentive,bevan2006measured}. The L1 incentive sweep in \S\ref{sec:l1} shows that this wedge becomes large and negative in exactly the intermediate-incentive interior region predicted by multitask theory.

\section{Strategic Policy-as-Code}\label{sec:mac}
\paragraph{Design desiderata.}
For regulated healthcare deployment, an admissible policy class $\Pi_A$ must be (i) \emph{inspectable} line-by-line for compliance review; (ii) \emph{regulable} over a fixed lever set whose semantics match real-world payer and hospital contracts, so that search cannot smuggle in new state variables, measurements, or distortion channels; (iii) \emph{sufficiently expressive} to admit state-conditional rules on observable aggregates (waiting, rejection, profit, utilization); and (iv) \emph{stress-testable} against the response class $\Pi_P$ of \S\ref{sec:formulation}. Black-box neural controllers \citep{zheng2022ai,dutting2024mechanism} satisfy (iii) but neither (i) nor (ii).

\paragraph{Policy class: typed executable programs.}
We instantiate $\Pi_A$ as a typed assignment-only DSL. A candidate policy bundle exposes the search-writable administrative expressions: $(\alpha,\beta)$, total and flexible capacities, the bonus pool and sharpness $\kappa$, the \kpi weight vector $(w_H,w_W,w_{\mathrm{rej}},w_C)$, and an optional \kpi-steering switch. The audit schedule is not a DSL field: the audit intensity $q_t$, the audit-hit function $p_{\mathrm{audit}}$, penalty multipliers, and clawback logic are fixed host-side configuration for a given rollout/evaluation setting. The bundle also includes selected provider-response expressions for effort, triage/resource requests, and coding candidate scoring, which instantiate the bounded-rationality response class $\Pi_P$.

L1 and L2 keep the provider-response rules and administrative rule programs fixed at the baseline of \S\ref{sec:produce}, except for the designated one-at-a-time diagnostic sweeps. L3 allows search over selected provider-response and administrative expression constants, but the simulator dynamics, patient generation, exposed features, metric computation, host-side clipping, feasibility projection, audit schedule, and settlement/audit routines remain fixed. Thus $\Pi_P$ remains a stress-test reference class for the simulator's response channels: L3 reweights bounded-rationality responses inside the same provider-response class rather than introducing a new provider model, new measurements, or a co-evolved opponent.

\paragraph{Search and evaluation.}
We instantiate this policy class with AlphaEvolve-style evolutionary code search \citep{novikov2025alphaevolve,romera2024funsearch}, implemented through \openevolve. A candidate program is first checked for syntactic and type validity, restricted to assignment-only edits over fixed policy fields, and then evaluated on short and full stochastic rollouts; Algorithm~\ref{alg:closed-loop} (App.~\ref{sec:app-algorithms}) summarizes the outer loop. The scalar fitness is the empirical estimator of the Stackelberg objective Eq.~\eqref{eq:stackelberg} with $\pi_P^*$ taken as the bounded-rationality best response, and the mixed objective uses a log-scaled funds-plus-reputation reward (Eqs.~\eqref{eq:fitness}--\eqref{eq:mix-reward}, App.~\ref{sec:app-formulation}). Programs that achieve high $G$ through up-coding or unsafe deferral incur a large $V$ and are demoted, so the search trajectory implements the safety constraint rather than relying on a post-hoc filter.

\section{Experiments}\label{sec:experiments}
The experiments follow a two-stage validation-to-discovery logic. We first verified that nine canonical healthcare stylized facts---\drg coding rent, profit-driven case-mix distortion, audit-induced channel substitution, target gaming, queueing capacity response, flexible-capacity allocation effects, quality-side multitasking, risk-adjustment up-coding, and bonus-pool misalignment---reproduce \emph{directionally} as the simulator's incentive, audit, and capacity levers are moved. The full validation map, with anchor citations and per-row signatures, is in Appendix Table~\ref{tab:external-validation-full}. With this external check in place, we use \medisim{} as a mechanism diagnostic across three layers: L1 maps provider responses over the incentive surface; L2 perturbs administrative levers and traces where provider pressure moves; L3 tests policy-as-code search under welfare, profit, and safety-penalized mixed objectives. All L1/L2 summaries report 30-seed means over horizon $T=200$ unless otherwise stated; Appendix~\ref{sec:app-l1-l2-setup} reports the incentive grid, L2 lever sweeps, defaults, and routing/steering switches.

\paragraph{Empirical findings.}
The experiments highlight four mechanism-level findings. First, classical healthcare failures occupy neighboring regions of an incentive phase diagram. Second, the balanced interior hides risk on less-visible margins: up-coding and rejection recede while delay and \kpi targeting intensify. Third, administrative controls move pressure across channels---audit closes coding but raises selection, bonus expansion worsens proxy alignment, and \kpi-steered flexible capacity raises waiting. Fourth, policy-as-code search follows its objective: pure-profit search amplifies coding, whereas the mixed objective reshapes incentive geometry and removes up-coding while preserving much of the return.

\subsection{L1: failure regimes form one incentive phase diagram}\label{sec:l1}
\begin{figure*}[t]
  \centering
  \includegraphics[width=0.7\textwidth]{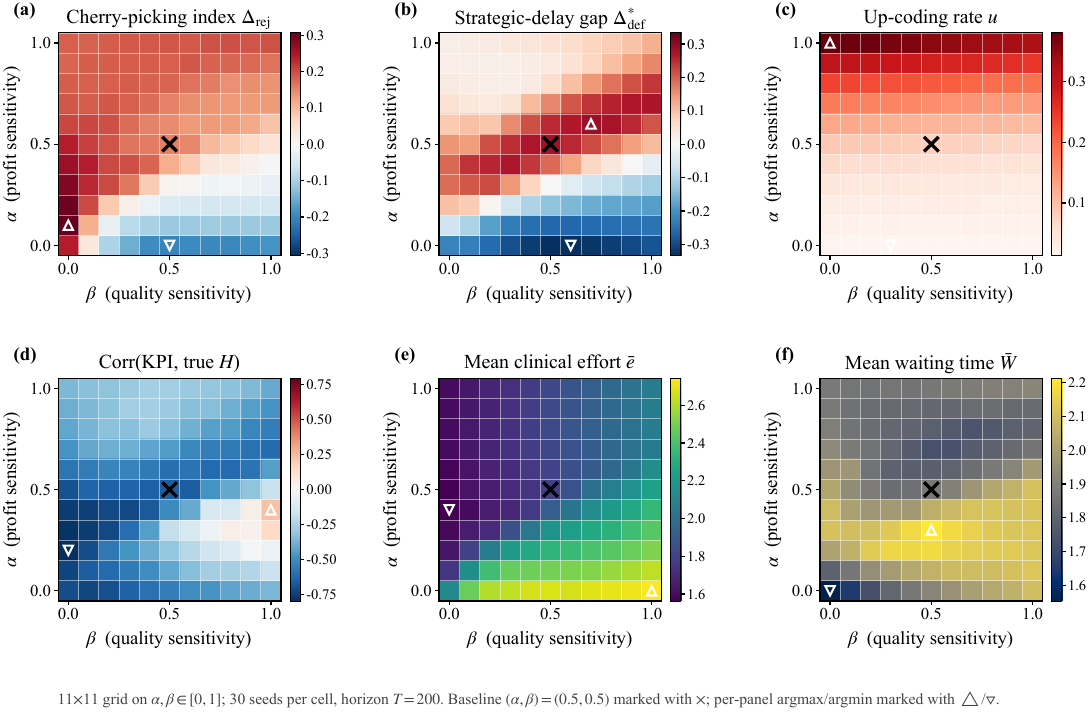}
  \caption{L1 incentive phase diagram. Each panel reports 30-seed means over the $11\times 11$ $(\alpha,\beta)$ grid. The grid separates low-incentive access rationing, profit-driven coding and selection, quality-driven effort and budget pressure, and balanced-interior delay and \kpi targeting.}
  \label{fig:l1phase}
\end{figure*}

L1 sweeps financial and quality sensitivities over an $11\times11$ grid. The response surface is structured but non-monotone: weak incentives ration access; profit pressure activates coding and selection (up-coding $0.226$, high-CMI rejection gap $0.182$); quality pressure suppresses these channels while increasing effort and solvency stress; and the balanced interior shifts pressure to delay and \kpi targeting (representative-regime metrics in Appendix Table~2). The profit- and quality-driven regions recover familiar incentive-theory predictions, but the balanced interior is most diagnostic: visible metrics improve while pressure moves into selective deferral and proxy targeting---high-CMI delay is $0.290$ versus $0.010$ for low-CMI patients, and measured \kpi becomes negatively aligned with true health ($-0.659$). L1 therefore exposes an interior failure hidden by single-channel diagnostics.

\subsection{L2: administrative levers move pressure across channels}\label{sec:l2}
\begin{figure*}[t]
  \centering
  \includegraphics[width=0.7\textwidth]{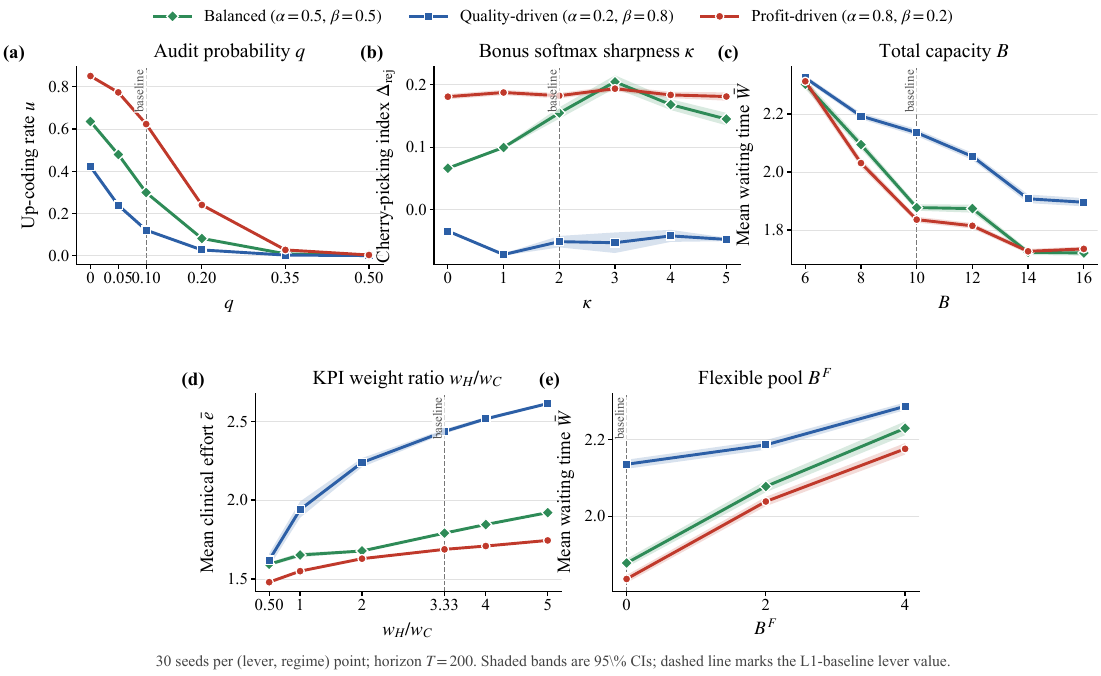}
  \caption{L2 one-at-a-time policy ablations. Curves report 30-seed means over horizon $T=200$ for balanced, quality-driven, and profit-driven regimes; shaded bands are 95\% confidence intervals. Audits suppress up-coding, capacity lowers waiting, and \kpi/bonus/flex levers induce nonlinear responses.}
  \label{fig:l2}
\end{figure*}

L2 reads administrative levers as pressure-tracing interventions: each lever is evaluated by its full response vector across coding, selection, delay, effort, and triage.

The audit sweep gives the cleanest substitution pattern. Raising audit probability lowers balanced-regime up-coding, while cherry-picking rises. Audit reallocates pressure from billing to selection, making access and delay diagnostics necessary alongside billing accuracy.

The bonus-pool sweep shows the measurement version of the same mechanism. Larger bonuses strengthen the reward attached to the measured \kpi. When that proxy is misaligned with true health, stronger incentives widen the wedge: at the endpoints of the balanced-regime sweep, \kpi--true-health correlation falls from $-0.447$ at $B^{\mathrm{pool}}=0$ to $-0.839$ at $B^{\mathrm{pool}}=15$. This is the main Goodhart-style L2 result.

The flexible-capacity sweep gives the operational version: under \kpi steering, balanced waiting rises from $1.88$ to $2.23$ as the flexible pool grows because additional capacity follows bonus-sensitive teams rather than the longest queues, while a steering-off diagnostic flattens the slope to $1.88\rightarrow1.88$ (App.~\ref{sec:app-l2}). Flexible capacity acts through its allocation rule, not through capacity volume alone.

\subsection{L3: search follows the incentives}\label{sec:l3}
L3 uses policy-as-code search to test how objective choice shapes the discovered mechanism family. Candidates are typed edits over the allowed DSL keys only (App.~\ref{sec:app-dsl}); implementation details and seed splits are in App.~\ref{sec:app-alphaevolve}.

\begin{table*}[t]
\centering
\footnotesize
\setlength{\tabcolsep}{4pt}
\renewcommand{\arraystretch}{1.02}
\begin{tabular}{@{}llrrrrr@{}}
\toprule
\textbf{Objective} &
\textbf{Method} &
\makecell[c]{\textbf{Fitness}$\uparrow$} &
\makecell[c]{\textbf{Wait}$\downarrow$} &
\makecell[c]{\textbf{Reject}$\downarrow$} &
\makecell[c]{\textbf{Upcoding}} &
\makecell[c]{\textbf{Funds}} \\
\midrule
Welfare & Greedy-Quality & 16.634 & 1.693 & 0.032 & 0.000 & 10.8 \\
Welfare & \lthreemethod{\alphaevolve} & \lthreebestcell{16.932} & \lthreebestcell{1.529} & \lthreebestcell{0.011} & 0.000 & 420.6 \\
\addlinespace[1pt]
Profit & Greedy-Profit & 121.846 & 1.784 & 0.068 & 0.758 & 7288.3 \\
Profit & \lthreemethod{\alphaevolve} & \lthreebestcell{122.046} & 1.786 & 0.068 & \lthreeriskcell{0.807} & \lthreebestcell{7353.7} \\
\addlinespace[1pt]
Mixed & Greedy-Profit & 13.580 & 1.784 & 0.068 & 0.758 & 7288.3 \\
Mixed & Best warm start & 13.607 & \lthreebestcell{1.674} & 0.058 & 0.000 & 5445.9 \\
Mixed & \lthreemethod{\alphaevolve} & \lthreebestcell{13.876} & 1.727 & \lthreebestcell{0.033} & 0.000 & 5480.7 \\
\bottomrule
\end{tabular}
\caption{L3 held-out performance. Entries are means over held-out test seeds. Blue cells identify searched policies; green cells mark key comparisons; red flags the profit-only up-coding risk.}
\label{tab:l3}
\end{table*}

The welfare objective is the positive-control case: \alphaevolve{} improves the welfare-family policy along every dimension in Table~\ref{tab:l3} while keeping up-coding at zero, and lifts doctor margin---search trims excessive effort cost while preserving the gains rewarded by the welfare objective (App.~\ref{sec:app-alphaevolve-method}). The profit objective is the reward-hacking diagnostic: \alphaevolve{} makes only a small refinement over \textsc{Greedy-Profit} and spends it through the coding channel (up-coding $0.758\!\to\!0.807$ alongside higher funds), showing that coding remains an available optimization channel when safety violations carry no penalty.The mixed objective produces the central L3 result. \alphaevolve{} keeps return close to the profit-oriented baseline, reduces the violation score from 8.170 to 3.002, halves rejection, and drives up-coding to zero. The mechanism-level change visible in App.~\ref{sec:app-alphaevolve-method} is that the searched policy lowers local bonus pressure while retaining aggregate performance: search therefore changes \emph{which channels} earn the return, rather than improving a scalar score.The warm-start ablation defines the scope of this result. With a neutral-only library, search fails to recover the mixed family. This scopes the policy-as-code interface as a structured refinement tool over meaningful policy priors.

\section{Discussion and Conclusion}\label{sec:conclusion}

\paragraph{Three classical episodes, one simulator.}
The episodes motivating \S\ref{sec:intro} map onto distinct $(\alpha,\beta)$ regions of L1's phase diagram---Dafny-style up-coding at the high-$\alpha$/low-$\beta$ corner, Silverman ownership-conditional case-mix distortion along $\alpha$ axis, and Bevan target gaming in the balanced interior. Recovering these patterns under one set of dynamics gives Medi-Sim its benchmark role: a common diagnostic environment for evaluating mechanisms under strategic provider response, rather than a collection of separately tuned stylized examples.

\paragraph{Channel substitution is the structural obstacle.}
L2 shows that closing one distortion channel can reopen another: audits suppress up-coding, but provider response shifts toward selection or delay, because $\Pi_P$'s five channels are coupled through Eq.~\eqref{eq:provider-util}. Evaluation must therefore track the whole response, not the single metric a mechanism was designed to improve.

\paragraph{Implications and closing.}
Medi-Sim makes three evaluation requirements explicit for healthcare policy ML: provider response should be endogenous, searched mechanisms should remain inspectable as code, and scalar reward should be decomposed into channel-level diagnostics so that reward gains through distortion are counted as failures. The released closed-loop environment instantiates these requirements with an Identify--Produce--Settle simulator and an \alphaevolve{}-style search interface over a typed DSL. In the held-out mixed-objective evaluation, the searched policy eliminates measured up-coding while retaining much of the profit-oriented baseline's funds, showing that policy-as-code search can reshape the channel through which return is earned. We hope \medisim{} makes provider-side strategic response a standard benchmark target for ML on healthcare policy.

\section*{Limitations}\label{sec:app-limitations}
\medisim is a mechanistic simulator, not a calibrated clinical deployment model, and the design trades external validity for per-channel inspectability in several deliberate ways.

\textit{Bounded-rationality response, not equilibrium.}
The response class $\Pi_P$ of \S\ref{sec:formulation} is implemented as closed-form behavioral rules driven by the local gradient of the team utility~\eqref{eq:provider-util} rather than as solved equilibria of an inner game. This is a tractable but strict approximation; how well it tracks a fully strategic provider population is an empirical question we do not resolve here.

\textit{Aggregated \drg and patient-choice routing.}
The default simulator uses an aggregated macro-\drg distribution and patient-choice registration. Active strategic hospital routing is disabled in L1/L2, and hospital \kpi steering acts only on the flexible capacity pool. These choices keep the selection and delay channels attributable to provider triage, which is essential for the L1/L2 attribution claims, but they limit the realism of routing-heavy interventions.

\textit{L3 depends on the warm-start library.}
As reported in \S\ref{sec:l3} and Appendix~\ref{sec:app-alphaevolve}, the L3 mixed-objective result is a refinement of a diverse warm-start library; with a neutral-only library, $K=200$ search does not recover the same family. We therefore present AlphaEvolve over \medisim as a feasibility demonstration of program search over the Mechanism-as-Code policy class rather than as a benchmark-winning algorithm.

\textit{Synthetic rollouts.}
Held-out evaluation uses fixed seed splits over synthetic rollouts; we do not establish real-world effectiveness, and any deployment use would require domain validation, calibration to local case mix, and safety, equity, and legal review.

\section*{Ethical Considerations}\label{sec:app-ethics}
The simulator studies high-stakes healthcare operations. Its policies must not be interpreted as clinical recommendations or administrative guidance for real hospitals without domain validation, safety review, and fairness analysis. Modeling behaviors such as up-coding, cherry-picking, and strategic delay is intended for detection and stress testing, not for operationalizing manipulation. Any future deployment would require safeguards for patient access, equity across case complexity, clinical safety, privacy, and legal compliance.

\bibliography{references}

\clearpage
\appendix

\section*{Appendix Contents}
\noindent\textbf{Anonymous code.}
The anonymized code and reproduction package is available at:
\begin{quote}
\small\url{https://anonymous.4open.science/r/Medi_Sim-CC85/}
\end{quote}

\noindent\textbf{Appendix guide.}
\begin{itemize}[leftmargin=*,topsep=2pt,itemsep=1pt,parsep=0pt]
  \item Appendix~\ref{sec:app-algorithms}: execution and search algorithms.
  \item Appendix~\ref{sec:app-formulation}: formal Stackelberg definitions.
  \item Appendix~\ref{sec:app-env}: environment functional forms.
  \item Appendix~\ref{sec:app-related}: related work.
  \item Appendix~\ref{sec:app-l1}: per-channel L1 phase-diagram anatomy.
  \item Appendices~\ref{sec:app-l2-levers}--\ref{sec:app-l2}: additional L2 diagnostics.
  \item Appendix~\ref{sec:app-l1-l2-setup}: L1/L2 setup and hyperparameters.
  \item Appendix~\ref{sec:app-external-validation}: external stylized-fact validation.
  \item Appendix~\ref{sec:app-alphaevolve}: L3 search diagnostics and discovered policies.
  \item Appendix~\ref{sec:app-dsl}: implementation details and DSL guardrails.
  \item Appendix~\ref{sec:app-terms-arrivals}: terminology and arrival-process note.
\end{itemize}

\setcounter{algorithm}{0}
\renewcommand{\thealgorithm}{A.\arabic{algorithm}}
\section{Execution and Search Algorithms}\label{sec:app-algorithms}
Algorithm~\ref{alg:sim-loop} expands the IPS diagram in Figure~\ref{fig:overview} into the period-by-period simulator loop. Algorithm~\ref{alg:closed-loop} gives the outer optimization loop used by L3.

\begin{algorithm}[t]
\caption{\medisim simulator execution loop}\label{alg:sim-loop}
\small
\begin{algorithmic}[1]
\Require Hospital capacity and queues, hospital provider teams, hospital administrator policy, reimbursement module, \kpi/bonus rule, initial state $X_1$.
\For{$t=1,\ldots,T$}
  \State \textbf{Hospital policy step:} the hospital administrator sets period levers $u_t$, including incentives, capacity, bonus-pool, \kpi weights, audit intensity, and optional routing or steering parameters.
  \State \textbf{Patient generation:} sample arrivals $\mathcal P_t$ from the configured arrival process and assign latent clinical attributes such as true group, urgency, tolerance, and true complexity $\mathrm{CMI}_i^\star$.
  \State \textbf{Identify / coding:} hospital coding staff map each arriving patient's clinical presentation to a billable group $\hat g_{i,t}$; the gap between true and coded complexity realizes the coding wedge.
  \State \textbf{Routing and queueing:} register or route patients to hospital care-team queues, carry over backlog, and increment waiting time for patients not served in the current period.
  \For{each hospital care team $j$}
    \State Observe local queue $\mathcal Q_{j,t}$, capacity, fatigue, incentives, and prior \kpi signals.
    \State Choose provider-side actions: accept/defer/reject, effort $E_{ij,t}$ for accepted cases, and resource request $R_{ij,t}$ under the response class $\Pi_P$.
  \EndFor
  \State \textbf{Capacity resolution and treatment:} enforce hard capacity constraints, treat selected patients, keep unserved patients queued or rejected according to the triage decision, and realize health output $H_{ij,t}$ and cost $C_{ij,t}$.
  \State \textbf{Settlement:} compute reimbursement from the reported billing group $\hat g_{i,t}$, aggregate margin and \kpi scores, apply audit penalties, and allocate bonuses.
  \State \textbf{State update:} update funds, reputation, queues, provider fatigue, \kpi history, and the next hospital state $X_{t+1}$.
\EndFor
\Ensure Rollout trajectory, hospital administrator return, and channel-level diagnostics.
\end{algorithmic}
\end{algorithm}

\begin{algorithm}[t]
\caption{Closed-loop policy search over \medisim rollouts}\label{alg:closed-loop}
\small
\begin{algorithmic}[1]
\Require Initial policy library $\mathcal L_0$, program-search operator, validation seeds, held-out test seeds, rollout horizon $T$.
\For{$k=1,\ldots,K$}
  \State \textbf{Deploy:} select or propose a typed hospital policy program $\pi_A^{(k)}$ over the allowed DSL fields.
  \State \textbf{Simulate:} evaluate $\pi_A^{(k)}$ by running Algorithm~\ref{alg:sim-loop} on stochastic validation rollouts.
  \State \textbf{Score:} compute scalar return $G$, safety/distortion penalty $V$, variance penalty, and diagnostic metrics.
  \State \textbf{Update:} retain or mutate policy programs according to validation fitness and the search operator.
\EndFor
\State Select the best validation candidate and evaluate it once on held-out seeds.
\Ensure Searched policy, held-out rollout profile, and search diagnostics.
\end{algorithmic}
\end{algorithm}

The closed loop contains two feedback channels. Within-rollout feedback is economic and operational: hospital incentives and routing rules shape coding, triage, treatment effort, congestion, settlement, and the next hospital state. Across-rollout feedback is learning-driven: the program-search operator updates the distribution of candidate hospital policies after observing rollout scores and diagnostics, which shifts the strategic and congestion regime explored by later rollouts.

\section{Formal Definitions for the Stackelberg Formulation}\label{sec:app-formulation}
This appendix expands the compact statement of \S\ref{sec:formulation}.

\paragraph{Hospital state.}
The state at time $t$ is
\begin{equation}\label{eq:hospital-state}
X_t = \big(F_t,\ Q_t,\ \{\mathcal Q_{j,t}\}_{j=1}^J,\ \mathrm{KPI}_{t-1},\ \mathrm{Rep}_t\big),
\end{equation}
where $F_t$ denotes funds, $Q_t$ summarizes systemwide congestion, $\mathcal Q_{j,t}$ is the queue registered to team $j$, $\mathrm{KPI}_{t-1}$ is the previous-period measured performance vector, and $\mathrm{Rep}_t$ is public reputation. The full per-patient state, the patient stream, and the audit signal are observable through these summaries and the exposed state features described in \S\ref{sec:env}.

\paragraph{Leader action.}
The hospital administrator's mechanism action is
\begin{equation}\label{eq:leader-action}
\begin{aligned}
u_t=\big(&\alpha_t,\beta_t,B^{\mathrm{tot}}_t,B^{\mathrm{flex}}_t,B^{\mathrm{pool}}_t,\\
&w_H,w_W,w_{\mathrm{rej}},w_C,\kappa,q_t,\xi_t\big),
\end{aligned}
\end{equation}
with the per-component semantics given in \S\ref{sec:formulation}. The flexible capacity pool $B^{\mathrm{flex}}_t$ is reallocatable across care teams, and \kpi steering through $\xi_t$ assigns that capacity using measured performance scores; see Appendix~\ref{sec:app-terms-arrivals}.

\paragraph{Safety/distortion penalty.}
The penalty term $V$ in Eq.~\eqref{eq:stackelberg} aggregates per-step penalties for unsafe waiting, excessive high-complexity deferral, up-coding, rejection, and insolvency; explicit weights are listed in App.~\ref{sec:app-dsl}.

\paragraph{Empirical fitness estimator.}
The search fitness used in \S\ref{sec:mac} is the empirical estimator of Eq.~\eqref{eq:stackelberg},
\begin{align}\label{eq:fitness}
\mathrm{Fitness}(\pi)=&\ \mathbb E_s[G(\pi;s)]-\lambda_{\mathrm{unsafe}}V(\pi)\nonumber\\
&-\lambda_{\mathrm{var}}\mathrm{Var}_s(G(\pi;s)),
\end{align}
where $G(\pi;s)\equiv G^o(\pi_A,\pi_P^*;s)$ instantiates the per-objective discounted return of Eq.~\eqref{eq:leader-return} with $\pi_P^*\in\Pi_P(\pi_A)$ taken as the bounded-rationality best response.

\paragraph{Mixed-objective per-step reward.}
For the mixed objective, one-step reward log-scales funds before combining them with reputation:
\begin{equation}\label{eq:mix-reward}
r_t^{\mathrm{mix}}=0.5\cdot\frac{\log(F_t)}{10}+0.5\cdot\mathrm{Rep}_t.
\end{equation}
This return enters the search fitness only and is not interpreted as a standalone welfare or economic-value measure.

\section{Environment Functional Forms}\label{sec:app-env}
This appendix collects the closed-form expressions referenced in \S\ref{sec:env}.

\paragraph{Arrival process.}
The default arrival process is homogeneous Poisson,
\begin{equation}
N_t = |\mathcal P_t| \sim \mathrm{Poisson}(\lambda_{\mathrm{arr}}),
\label{eq:arrival-poisson}
\end{equation}
with per-patient attributes drawn as described in \S\ref{sec:identify}. Appendix~\ref{sec:app-terms-arrivals} states when non-Poisson kernels can be substituted.

\paragraph{Coding candidate score.}
For candidate reported group $g$, let $\Delta R_i(g)$ be incremental reimbursement and let $\Delta c_i(g)$ be coded-complexity inflation. The candidate score is
\begin{equation}\label{eq:coding-score}
\begin{aligned}
s_i(g)=&\ \alpha_t\,\gamma_{\mathrm{code}}\,\Delta R_i(g)\\
&-p_{\mathrm{audit}}(\Delta c_i(g))\,\lambda_{\mathrm{pen}}\,[\Delta R_i(g)]_+\\
&-r_0\,\eta_{\mathrm{eth}}\,[\Delta c_i(g)]_+,
\end{aligned}
\end{equation}
where $p_{\mathrm{audit}}(\cdot; q_t)$ is the host-defined audit-hit function under the configured audit intensity $q_t$. The coded group is chosen by the configured score-based rule: a softmax over $\{s_i(g)\}$ in stochastic response mode, and its zero-temperature $\arg\max_g s_i(g)$ variant in deterministic L3 search/evaluation. Both variants use the same candidate score; intermediate aggregate up-coding arises from patient heterogeneity and incentive-dependent score comparisons rather than from a separate coding mechanism.

\paragraph{Treatment production and cost.}
For an accepted patient $i$ treated by unit $j$, true health output follows a diminishing-return production,
\begin{equation}\label{eq:treatment}
H_{ij,t}=\mathrm{Skill}_j\left(1-\exp\!\left[-\lambda\,\frac{E_{ij,t}}{\mathrm{CMI}_i^\star+\epsilon}\right]\right)R_{ij,t},
\end{equation}
with effort $E_{ij,t}\ge 0$ and resource indicator $R_{ij,t}\in\{0,1\}$. Cost is convex in effort,
\begin{equation}\label{eq:cost}
C_{ij,t}=C_{\mathrm{fixed}}\,R_{ij,t}+\omega\,E_{ij,t}^{\phi}, \qquad \phi>1.
\end{equation}

\paragraph{Bonus tournament and marginal pressure.}
Bonuses are allocated through a softmax tournament with sharpness $\kappa$ and pool $B^{\mathrm{pool}}_t$,
\begin{align}
s_{j,t}&=\frac{\exp(\kappa\,\mathrm{KPI}_{j,t})}{\sum_{\ell=1}^J\exp(\kappa\,\mathrm{KPI}_{\ell,t})},\label{eq:bonus-softmax}\\
\mathrm{Bonus}_{j,t}&=B^{\mathrm{pool}}_t\,s_{j,t},
\end{align}
with local marginal bonus pressure
\begin{equation}\label{eq:bonus-pressure}
\frac{\partial \mathrm{Bonus}_{j,t}}{\partial \mathrm{KPI}_{j,t}}=B^{\mathrm{pool}}_t\,\kappa\,s_{j,t}(1-s_{j,t}),
\end{equation}
which feeds back into effort, triage, and \kpi-targeting behavior.

\section{Related Work}\label{sec:app-related}

\paragraph{AI and Reinforcement Learning for Clinical Decision-Making.}
A substantial line of work treats healthcare as a sequential decision problem at the patient level, learning treatment policies from electronic health records under offline RL formulations. 
The AI Clinician learns vasopressor and fluid policies for sepsis from MIMIC-III and reports lower in-hospital mortality relative to observed clinician behavior \citep{komorowski2018ai}, and broader surveys catalogue similar formulations for chronic disease management, anesthesia, and ventilation \citep{yu2021reinforcement}. 
Methodological guidelines emphasize off-policy evaluation, distributional shift, and reward specification as the core obstacles to safe clinical deployment \citep{gottesman2019guidelines}. 
This literature optimizes the \emph{clinician's} action on a single patient and assumes a fixed reward and a fixed environment. 
\medisim{} operates one level above this layer: the action is a hospital administrative rule (incentive coefficients, audit probability, capacity allocation, hospital \kpi weights), the environment is a population of strategic providers, and the realized policy is the composition of the hospital administrator's mechanism with the providers' best response. 
Patient-level clinical RL is thus complementary to, not a substitute for, the mechanism-level question we study.

\paragraph{Healthcare Incentives and Operations.}
Provider behavior under payment rules has been studied for three decades through agency theory and operational flow. 
Cost and quality incentives under prospective payment can produce creaming, skimping, dumping, and intensity shifts \citep{ellis1998creaming, ma1994health, eggleston2005multitasking}. 
Empirical studies on Diagnosis-Related Groups confirm that administrative pricing shifts directly alter case mix and coding \citep{dafny2005hospitals, silverman2004medicare}, and more recent evidence from Medicare Advantage shows that risk-adjusted diagnosis-based reimbursement raises reported risk scores by 6--16\% without commensurate change in underlying morbidity \citep{geruso2020upcoding}. 
On the operational side, hospital flow is highly sensitive to capacity: queueing analyses show wait times escalate non-linearly near full utilization \citep{green2002beds, green2006queueing, cayirli2003outpatient}, and flexible bed allocations partially mitigate the load under realistic routing constraints \citep{bekker2017flexible}. 
Healthcare-specific simulation has been used for decades to study these flows, but predominantly as isolated discrete-event models without strategic agents \citep{brailsford2009simulation}. 
\medisim{} integrates this macroeconomic payment layer and microeconomic operational layer into a single executable closed loop in which coding, capacity, and revenue interact.

\paragraph{Measurement, Auditing, and Strategic Gaming.}
Prospective payment introduces an informational wedge between ground-truth clinical complexity and reported billing groups.
Pervasive up-coding has been documented across ownership structures \citep{silverman2004medicare, geruso2020upcoding}, and economic audit theory characterizes how monitoring rates and penalties govern this ``DRG creep'' \citep{kuhn2008upcoding}. 
Performance measurement creates a parallel friction: when scored metrics are imperfect proxies for the planner's objective, high-powered incentives intensify multitasking distortions and target gaming \citep{holmstrom1991multitask, baker1992incentive, bevan2006measured}. 
Systematic reviews of pay-for-performance in healthcare reach a consistent verdict that contract design and local context determine whether bonuses improve care or merely move metrics \citep{eijkenaar2013p4p, vanherck2010p4p}. 
Formal taxonomies of Goodhart-style failures (regressional, extremal, causal, adversarial) clarify why such gaming is structural \citep{manheim2018goodhart}. 
\medisim{} models the hospital coding wedge and the hospital measurement wedge as dynamic, interactive processes, attributing each distortion to the agent that actually makes the choice (clinician/hospital coder for up-coding, triage for selection and delay, hospital administrator for routing and incentive design), so that channel-level distortion becomes a measurable property of the rollout.

\paragraph{Strategic Machine Learning and Algorithmic Mechanism Design.}
Our problem interfaces directly with the machine-learning literature on agents that respond to a deployed rule. \emph{Strategic classification} formalizes Stackelberg learning against a manipulable agent \citep{hardt2016strategic} and has been extended to revealed-preference observations \citep{dong2018strategic} and to practical, end-to-end differentiable training pipelines \citep{levanon2021strategic}. 
\emph{Performative prediction} generalizes this to settings in which the deployed model itself shifts the data distribution and characterizes performative equilibria \citep{perdomo2020performative}. 
The reward-shaping side of this literature gives formal notions of \emph{reward hacking}: optimizing an imperfect proxy can never be made safe by narrowing the reward function alone under mild structural assumptions \citep{skalse2022reward}. 
In parallel, differentiable economics has used deep architectures to construct optimal multi-dimensional auctions from data \citep{dutting2024mechanism}. 
These neural mechanisms achieve strong objective values but yield black-box controllers whose parameters resist verification. 
\medisim{} adopts the strategic-learning stance (provider best response is part of the environment) and the algorithmic-mechanism-design objective (synthesize a rule that is robust to that response), but constrains the policy class to read-and-comply programs, which is what regulated healthcare deployment actually demands.

\paragraph{Multi-Agent Simulations and Program-Space Policy Search.}
Designing macro-policies by modeling micro-agent adaptation has recently scaled through multi-agent reinforcement learning and large-language-model agents. 
The AI Economist frames optimal taxation as a two-level RL problem in which agents and a planner co-adapt \citep{zheng2022ai}. 
LLM-based generative agents extend this paradigm to believable populations of social actors \citep{park2023generative}, and frameworks such as Concordia operationalize generative agent-based modeling with grounded actions and an explicit game-master adjudication layer \citep{vezhnevets2023concordia}. 
The LLM-economist line scales mechanism design to large generative simulacra of taxpayers and consumers \citep{karten2025llmeconomistlargepopulation}. 
However, these systems instantiate consumption, taxation, or social interaction as their domain primitives, not capacity-constrained clinical care, and they typically train neural controllers whose decision logic is not directly auditable. 
To navigate a non-differentiable and tightly regulated mechanism space without sacrificing inspectability, we build on LLM-guided evolutionary program search: 
FunSearch surpassed best-known constructions in extremal combinatorics and online bin packing by treating an LLM as a structured mutation operator over code \citep{romera2024funsearch}; 
ELM showed that LLM-based mutation can drive open-ended evolution in code-defined domains \citep{lehman2022elm}; 
AlphaEvolve generalizes this to algorithmic discovery across mathematics, hardware, and learning systems \citep{novikov2025alphaevolve}; 
and Eureka demonstrates that the same recipe can author RL reward functions that outperform expert humans on a majority of robotics tasks \citep{ma2024eureka}. 
More broadly, LLMs have been studied as black-box optimizers in their own right \citep{yang2023large}. 
\medisim{} adapts this paradigm to healthcare mechanism design: 
a candidate hospital policy is a typed executable program over a constrained hospital policy DSL of administrative levers, and the evaluator scores it on stochastic multi-agent rollouts under safety, access, and distortion constraints, yielding policies that are simultaneously high-performing, robust to provider best response, and human-readable.

\section{A Per-Channel Anatomy of the L1 Phase Diagram}\label{sec:app-l1}
The headline phase diagram of \S\ref{sec:l1} reports six top-line outcomes on the $11\times 11$ $(\alpha,\beta)$ grid. Each of those outcomes, however, is a slice through the joint response of five behavioral channels, and reading any single slice in isolation can mislead. The detailed panels in this appendix, including access decomposition, strategic-delay decomposition, clinical budget panels, and team-level \kpi panels, decompose each headline outcome into its constituent primitives. Two structural facts emerge that are not visible at the headline resolution.

First, the four-regime story is preserved at higher resolution: every additional panel either localizes a known regime more sharply or reveals a previously hidden margin without contradicting the headline.

Second, the channels are arranged in a \emph{substitution lattice}: in regions where one distortion saturates against a structural floor or ceiling, an adjacent channel becomes active to relieve the same shadow price. This lattice is the structural reason the headline regimes are separable and is the strongest piece of evidence we have that the IPS decomposition of \S\ref{sec:env} is the right factorization for \medisim.

\begin{table}[t]
\centering
\small
\resizebox{\columnwidth}{!}{%
\begin{tabular}{lrrrrrrrr}
\toprule
Regime & $\alpha$ & $\beta$ & $\Delta$ rej. & High rej. & Low rej. & Upcode & Effort & Funds \\
\midrule
Low-incentive & 0.0 & 0.0 & \tabrisk{0.241} & \tabrisk{0.542} & \tabrisk{0.301} & 0.014 & 2.128 & \tabrisk{23.4} \\
Profit-driven & 0.8 & 0.2 & 0.182 & 0.182 & 0.000 & \tabrisk{0.226} & 1.688 & \tabgood{5451.7} \\
Quality-driven & 0.2 & 0.8 & -0.051 & 0.026 & 0.078 & 0.027 & \tabrisk{2.435} & \tabrisk{15.2} \\
Balanced & 0.5 & 0.5 & 0.155 & 0.157 & 0.002 & 0.080 & 1.790 & 3486.6 \\
\bottomrule
\end{tabular}%
}
\caption{Representative L1 regimes. Entries are 30-seed means over horizon $T=200$; $\Delta$ rej. denotes high-\cmi minus low-\cmi rejection. Colored cells mark regime-defining extrema among the four rows.}
\label{tab:l1}
\end{table}

\begin{figure*}[!t]
  \centering
  \includegraphics[width=0.96\textwidth]{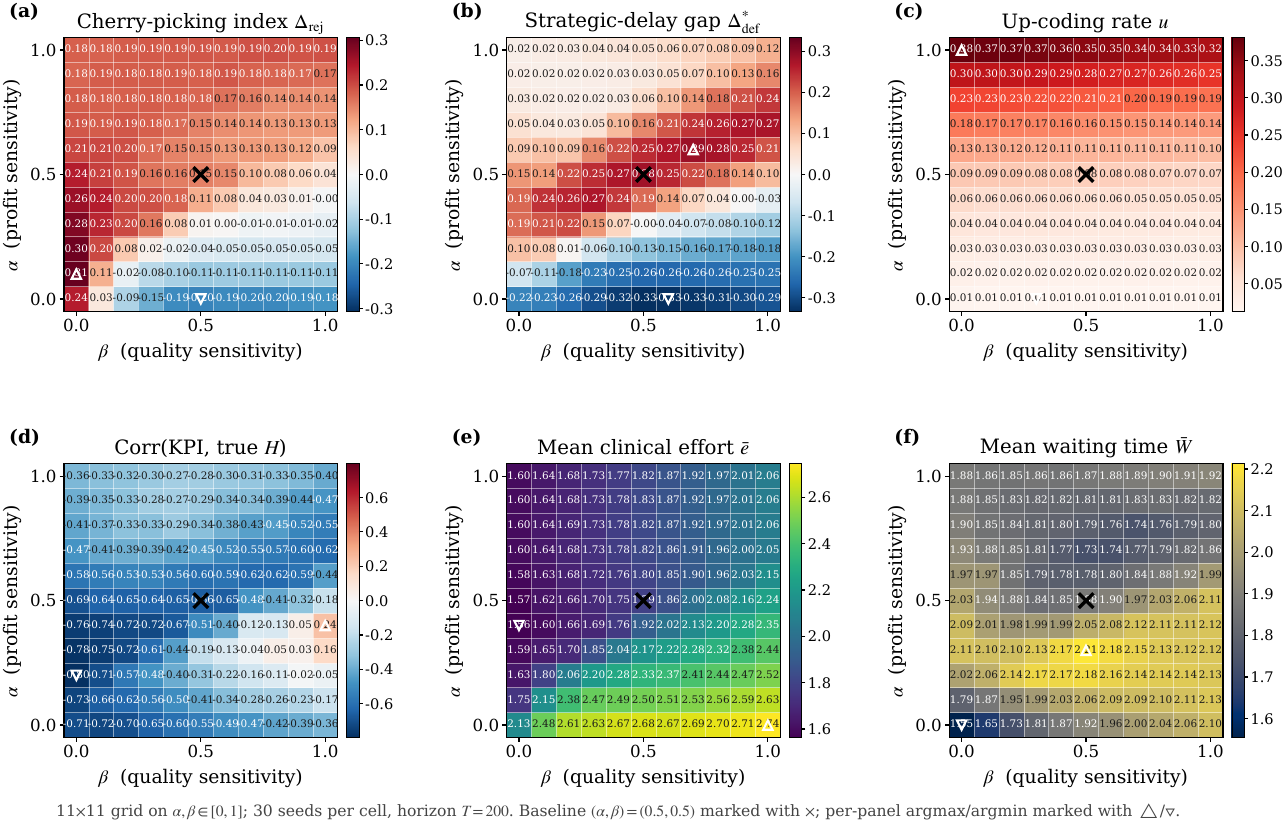}
  \caption{Cell-annotated L1 phase diagram. Each cell reports the 30-seed mean of the indicated outcome on the $11\times 11$ $(\alpha,\beta)$ grid at horizon $T=200$; $\times$ marks the balanced baseline and $\triangle$/$\triangledown$ mark the per-panel argmax/argmin. The numerical annotations make it possible to verify the regime boundaries discussed in \S\ref{sec:app-l1-access}--\S\ref{sec:app-l1-kpi} cell by cell.}
  \label{fig:l1annotated}
\end{figure*}

\subsection{Reading the annotated phase diagram}\label{sec:app-l1-annotated}
This subsection provides cell-level annotations (Figure~\ref{fig:l1annotated}) for the L1 grid (Figure~\ref{fig:l1phase}), serving as the reference for the detailed decompositions that follow. Three patterns are immediately visible.
\textit{(i)} The cherry-picking index $\Delta_{\mathrm{rej}}$ (panel a) attains its maximum at $(\alpha,\beta)=(0.1,0.0)$ with value $0.31$ and inverts in the low-$\alpha$/high-$\beta$ band, reaching $-0.20$ at $(0.0,0.5)$. The transition generally weakens as quality weight rises, but it is not strictly monotone cell by cell.
\textit{(ii)} The strategic-delay gap $\Delta^{*}_{\mathrm{def}}$ (panel b) is \emph{not} monotone in either variable: its maximum is $0.29$ at $(0.6,0.7)$, well inside the mixed-incentive corridor, and the panel contains a negative stripe at high $\beta$, low $\alpha$ where deferral protects \emph{low}-complexity cases. This pattern is decomposed in the dedicated delay analysis of \S\ref{sec:app-l1-delay}.
\textit{(iii)} The up-coding rate $u$ (panel c) is monotone in $\alpha$ at every $\beta$: values run from about $0.01$ at low $\alpha$ to $0.38$ at $\alpha=1.0$. Variation along $\beta$ at fixed $\alpha$ is much smaller but not zero, reaching about $0.06$ in the highest-$\alpha$ row. Coding is therefore the most nearly one-dimensional channel in the parameter pair $(\alpha,\beta)$, a fact that matters when we discuss the audit-coding-selection substitution in \S\ref{sec:app-l1-kpi}.
The remaining panels, $\mathrm{Corr}(\mathrm{KPI},H)$, mean effort, and mean waiting, are each interpreted in the dedicated subsections below.

\subsection{The access channel: rejection is only half the story}\label{sec:app-l1-access}
\begin{figure*}[!t]
  \centering
  \includegraphics[width=0.96\textwidth]{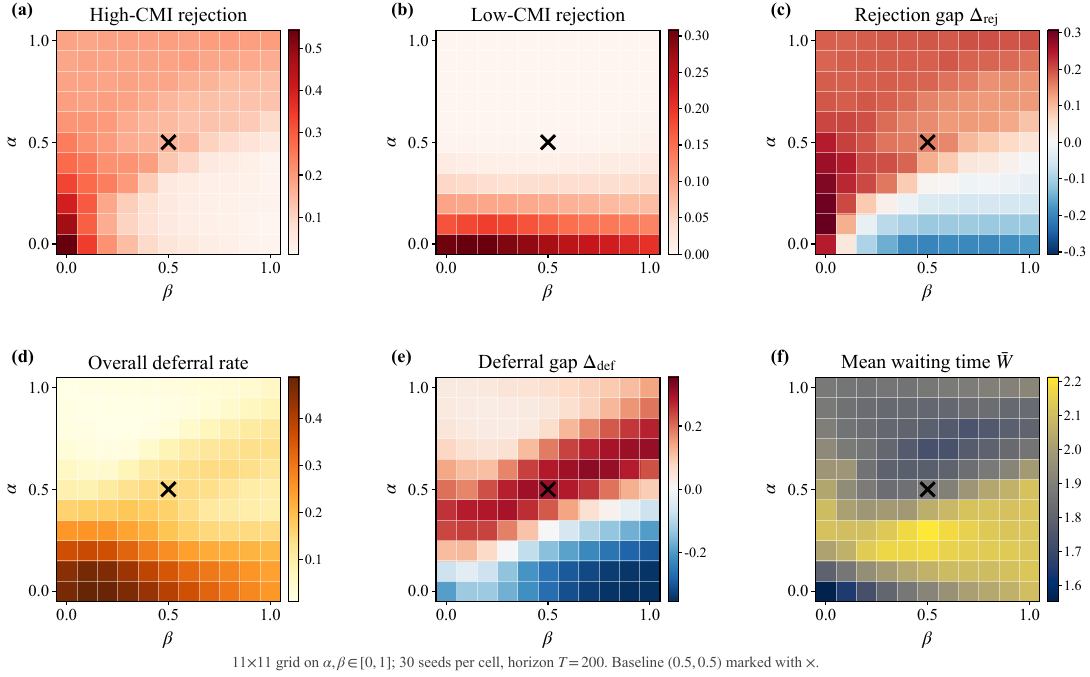}
  \caption{Access decomposition on the L1 grid. Panels (a) and (b) split rejection by true complexity into high-CMI and low-CMI components, panel (c) is the resulting cherry-picking gap, panels (d) and (e) do the same decomposition for deferral, and panel (f) reproduces the realized mean waiting time. The contrast between (a,b) and (d,e) is the central content of this appendix: as $\alpha$ rises, the access burden migrates from rejection to deferral, but only for high-complexity patients.}
  \label{fig:l1access}
\end{figure*}

Cherry-picking is conventionally measured as the gap between high-complexity and low-complexity rejection rates, and on that summary statistic alone (Figure~\ref{fig:l1annotated}(a)) the profit-driven corner looks unambiguous. The decomposition in Figure~\ref{fig:l1access} reveals that this headline hides a subtler structural fact.

At low $\alpha$, both rejection rates are high in absolute terms. The providers lack a positive margin and treat acceptance as a pure fatigue cost, so the gap between them is small (Figure~\ref{fig:l1access}, panels a--b). As $\alpha$ rises into the profit corner, both rates fall because throughput becomes valuable, but low-CMI rejection falls faster: cheap cases are unconditionally profitable. The gap therefore widens mainly because they accept simple ones more eagerly. The conventional cherry-picking index captures the \emph{size} of the gap but misattributes its \emph{source}.

The deferral panels (d) and (e) tell the complementary story. Overall deferral is not monotone in $\alpha$; the more stable signal is the high-CMI minus low-CMI deferral gap. That gap becomes positive across the moderate- and high-$\alpha$ band, reaching 0.316 at $(0.6,0.7)$, while low-$\alpha$/high-$\beta$ cells reverse sign. This is a substitution pattern: when outright rejection becomes costly, providers shift cost-shedding into the delay channel, targeting the same patient subpopulation. Panel (f) confirms that this substitution is not free: realized waiting peaks at $(\alpha,\beta)=(0.3,0.5)$, in the same moderate-incentive corridor where access pressure is redistributed across rejection and deferral.

The headline rejection gap heatmap suggests that the profit corner is the single worst access regime. Figure~\ref{fig:l1access} reveals two distinct worst-case access regimes. Between them, a moderate-incentive corridor produces queue-mediated congestion as rejection and deferral trade off against each other. This third failure mode is invisible to audit-only interventions.This micro-level structure underlies the channel substitution results of \S\ref{sec:l2}.

\subsection{Strategic delay as a cost-relief substitute}\label{sec:app-l1-delay}
\begin{figure*}[!t]
  \centering
  \includegraphics[width=0.96\textwidth]{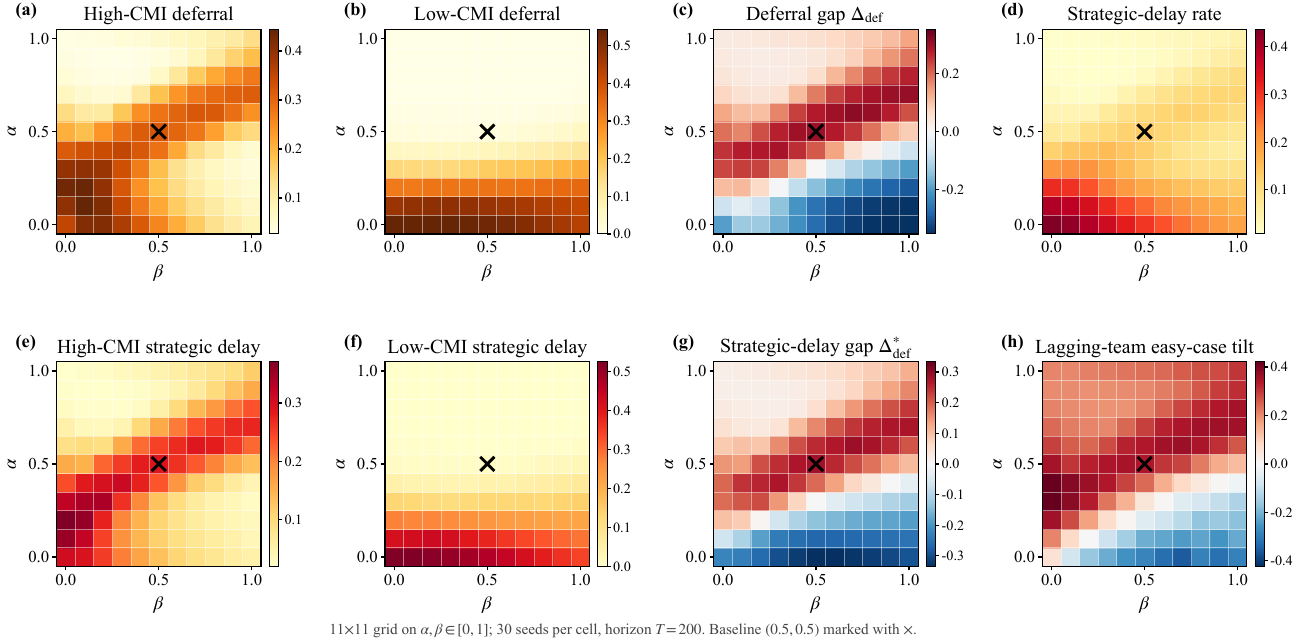}
  \caption{Strategic-delay decomposition on the L1 grid. Panels (a) and (b) report deferral rates separately for high-CMI and low-CMI patients; panel (c) is the deferral gap. Panels (d)--(f) restrict the deferral sample to delays that improve next-period team utility under Eq.~\eqref{eq:provider-util} (``strategic delay''); panel (g) is the strategic-delay gap; panel (h) is the easy-case CMI tilt of KPI-lagging teams.}
  \label{fig:l1delay}
\end{figure*}

Figure~\ref{fig:l1delay} isolates the delay channel from queue-mediated waiting and from involuntary deferral. The four-panel block (d)--(g) restricts the deferral sample to cases where the doctor chooses to defer \emph{despite available local bed capacity}; this is the operationalization of ``strategic delay'' used in the main text. Three findings emerge.

First, strategic delay is not concentrated on high-complexity patients everywhere. At low $\alpha$, low-CMI strategic delay can exceed high-CMI strategic delay, producing the negative gaps in panel~(g). In the mixed-incentive corridor the sign flips: high-CMI strategic delay dominates, consistent with the cost-relief motive in the microfoundation.

Second, the strategic-delay gap~(g) is the most clearly interior pattern on the grid. Its maximum $0.29$ sits at $(0.6, 0.7)$, where neither profit-corner rejection nor quality-corner effort fully dominates and delay becomes the cheapest cost-relief lever. Its minimum $-0.33$ sits at $(0.0, 0.6)$, where teams protect low-complexity throughput against the cost weight in $\mathrm{KPI}$. Neither pattern is observable from the headline rejection heatmap alone.

Third, the lagging-team easy-case tilt~(h) ranges from about $-0.35$ to $+0.42$ and captures a different mechanism from the strategic-delay gap. The tilt is a \emph{team-level} Goodhart signature: teams trailing in the KPI tournament shift toward easier cases to recover bonus share. This mode becomes salient when bonus pressure $B^{\mathrm{pool}} \kappa\, s_j(1 - s_j)$ is large (see Figure~\ref{fig:bonus} for the corresponding sensitivity).

Taken together, panels~(g) and~(h) furnish quantitative evidence for the multitasking claim in \S\ref{sec:l1}: where the headline KPI--health correlation is most negative, the responsible channels are not effort distortion but patient-level delay and team-level case-mix tilt acting in concert.

\subsection{Clinical-budget frontier: bounded health, unbounded cost}\label{sec:app-l1-clinical}
\begin{figure*}[!t]
  \centering
  \includegraphics[width=0.96\textwidth]{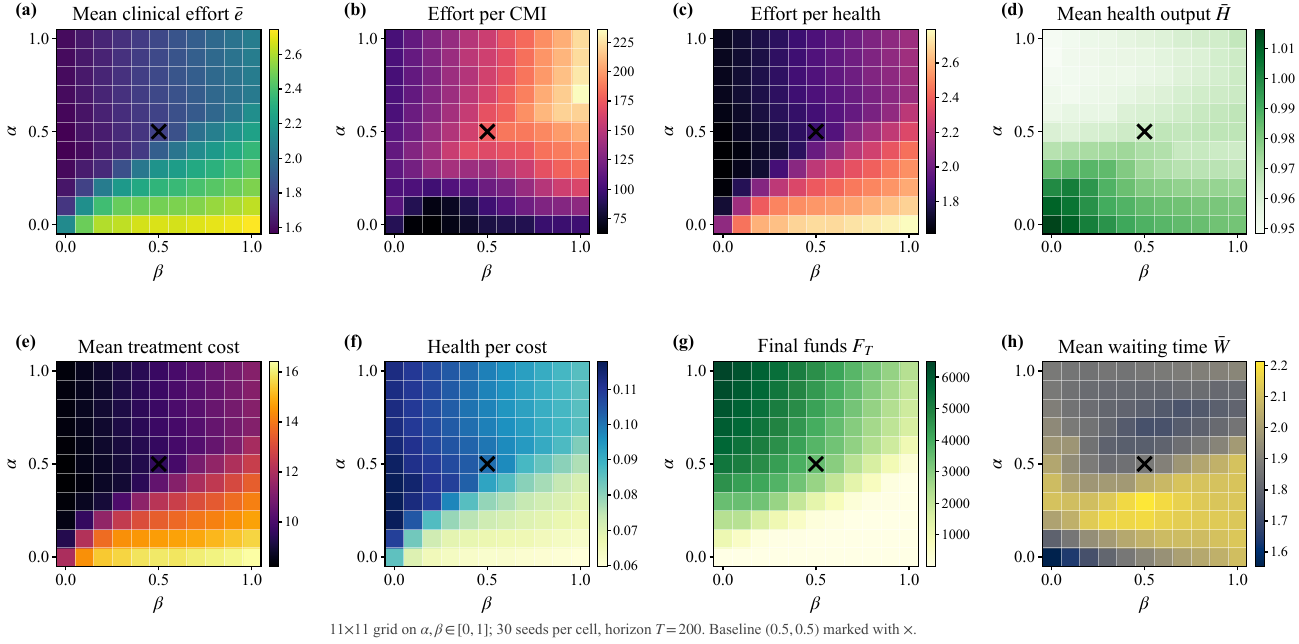}
  \caption{Clinical and budgetary outcomes on the L1 grid. Top row: mean clinical effort, effort per unit true CMI, and effort per unit realized health. Middle row: mean realized health, mean treatment cost, and health per cost. Bottom row: terminal funds $F_T$ and mean waiting time. The contrast between the narrow range of health (0.95--1.02) and the multi-order-of-magnitude range of terminal funds (0.4--6490) is the structural reason why interpreting health in isolation is misleading on this grid.}
  \label{fig:l1clinical}
\end{figure*}

The dominant qualitative feature of Figure~\ref{fig:l1clinical} is a scale asymmetry between clinical gains and financial exposure. Mean realized health (panel d) varies only in the narrow band $[0.95,1.02]$, while terminal funds (panel g) span more than four orders of magnitude. Mean effort (panel a) ranges from $1.56$ to $2.74$, mean cost (panel e) from $8.23$ to $16.41$, and health-per-cost efficiency (panel f) collapses from about $0.12$ to $0.06$ as $\beta$ rises. The diminishing return treatment production function of \S\ref{sec:produce} guarantees that health saturates while cost is convex in effort. The quality-driven corner therefore becomes insolvent  because it keeps buying effort on a saturated-health, high-cost segment of the production curve.

Another two secondary patterns are important for interpretation. The effort-per-CMI panel (b) reveals that gold-plating intensifies along $\beta$ even when the true clinical need is held constant in the denominator, ruling out a pure case-mix explanation. The effort-per-health panel (c) is essentially a mirror image of the health-per-cost panel (f), with the small but real implication that the marginal product of effort drops by a factor of $1.5\text{--}2\times$ as we cross the saturating region; a mechanism designer that uses health as a target without dividing by cost will keep paying for an effort margin that produces almost no measurable health benefit. The waiting panel (h) reproduces the U-shape already seen in Figure~\ref{fig:l1annotated}(f) but at higher resolution: waiting peaks at $2.21$ near $(\alpha,\beta)=(0.3,0.5)$, where selection has not thinned the queue fully.

\subsection{KPI proxy management at the team level}\label{sec:app-l1-kpi}
\begin{figure*}[!t]
  \centering
  \includegraphics[width=0.96\textwidth]{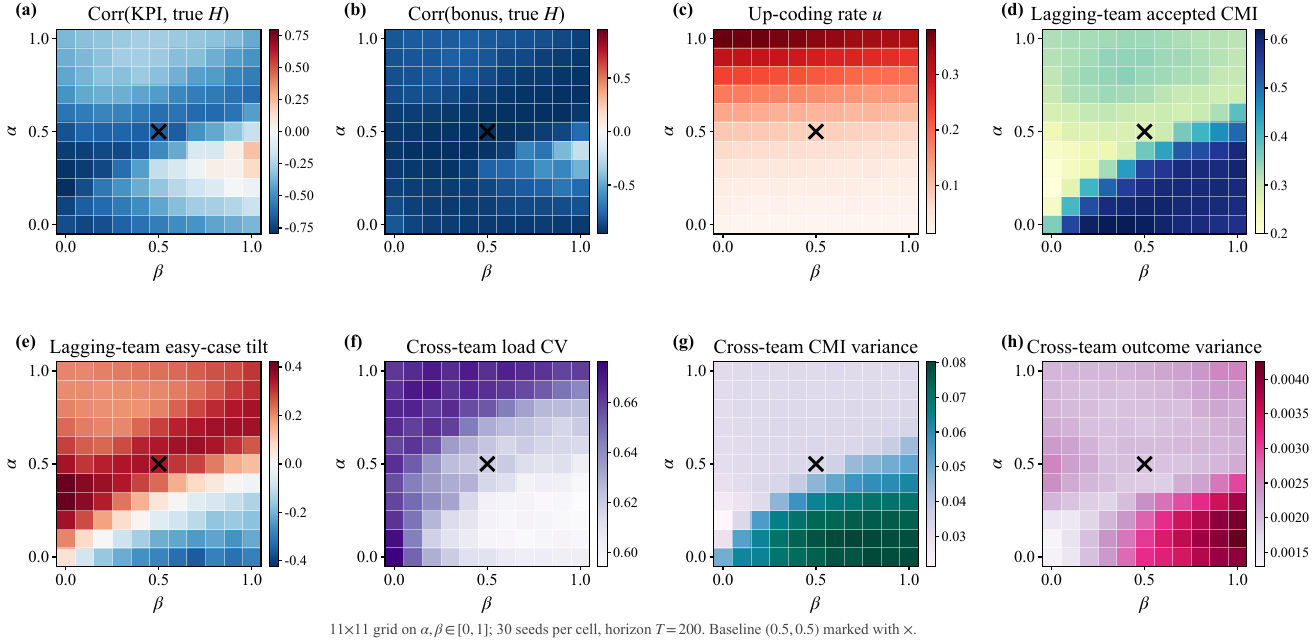}
  \caption{Team-level and KPI-level diagnostics on the L1 grid. Panels (a) and (b) report Pearson correlations of measured KPI and realized bonus against true health, on the cross-team panel. Panels (c)--(e) report the up-coding rate, mean accepted CMI by lagging teams, and the easy-case tilt of lagging teams. Panels (f)--(h) report cross-team coefficient of variation of load, of accepted CMI, and of clinical outcome.}
  \label{fig:l1team}
\end{figure*}

Figure~\ref{fig:l1team} is the team-level counterpart to the patient-level decomposition above. Two panels carry most of the interpretive weight.

Panel (a) plots $\mathrm{Corr}(\mathrm{KPI}_{j,t},\,H_{j,t})$ across teams and time on each $(\alpha,\beta)$ cell. The correlation is sharply negative in the low-$\beta$ band with low or moderate $\alpha$ (reaching $-0.80$ at $(0.2,0.0)$), is close to zero along a diagonal corridor through the interior, and becomes positive only in a high-$\beta$ corridor (peaking at $+0.24$ at $(0.4,1.0)$). The bonus-health correlation in panel (b) repeats this pattern with sharper amplitude because the softmax tournament amplifies any cross-team ordering present in $\mathrm{KPI}$.

Panel (d), the mean accepted CMI of KPI-lagging teams, can be interpreted as conditional workload level. It ranges from about $0.20$ to $0.62$ and records the case mix that lagging teams actually end up carrying. Panel (e) reports the lagging-team easy-case tilt, an acceptance-rate gap between low- and high-CMI patients that spans roughly $[-0.35,+0.42]$. This distinction matters in the low-$\alpha$/high-$\beta$ corner: lagging teams have high accepted CMI there because quality pressure protects clinically complex patients while weak profit pressure removes the incentive to shed them. The negative tilt in the same region confirms that these teams are not padding KPI with easy cases, but carrying a heavier high-CMI clinical burden. Conversely, positive values in panel (e) identify KPI-driven easy-case selection among lagging teams, a team-level pattern that complements the aggregate rejection-gap view in Figure~\ref{fig:l1annotated}(a).

Panels (f)--(h) are auxiliary diagnostics. We report cross-team dispersion in load, accepted case mix, and outcomes to check whether the patient-level distortions documented above also leave a team-level footprint. The clearest signal is the accepted-CMI variance, which rises in regions where KPI-lagging teams tilt toward easier cases. The bottom row also separates two qualitatively different dispersion regimes. In the low-$\beta$ band, weak quality pressure makes case-mix sorting the dominant source of cross-team heterogeneity: teams differ mainly in which patients they accept. In the low-$\alpha$/high-$\beta$ corner, by contrast, profit-driven shedding is weak and quality pressure protects high-CMI patients, so cross-team dispersion reflects uneven clinical burden, effort intensity, and outcome variation under a saturated production function.

\section{L2 Additional Lever Diagnostics}\label{sec:app-l2-levers}
The main text reports the three L2 mechanism findings. The remaining one-at-a-time sweeps serve as sanity checks and mechanism decompositions. Increasing total capacity from 6 to 16 reduces mean waiting in every regime, consistent with high-utilization queueing predictions. Raising the \kpi health-to-cost weight ratio increases clinical effort while weakening budget sustainability. The bonus-sharpness sweep is non-monotone because the local derivative $B^{\mathrm{pool}}\kappa s_j(1-s_j)$ is weak when the tournament is flat and also weak when shares saturate.

\begin{figure*}[t]
  \centering
  \includegraphics[width=0.65\textwidth]{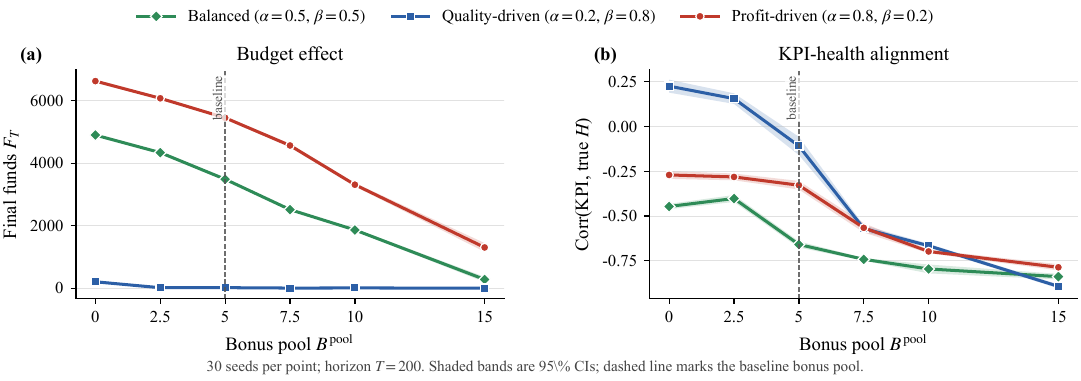}
  \caption{L2 bonus-pool ablation. Points are 30-seed means over horizon $T=200$; shaded bands are 95\% confidence intervals and the dashed line marks the baseline pool. Larger pools reduce funds and can weaken KPI-health alignment.}
  \label{fig:bonus}
\end{figure*}

Figure~\ref{fig:bonus} isolates the bonus-pool channel behind the main-text Goodhart result. Larger pools reduce final funds in all three regimes. In the balanced regime, \kpi--true-health correlation falls from $-0.447$ at $B^{\mathrm{pool}}=0$ to $-0.839$ at $B^{\mathrm{pool}}=15$, showing that stronger measured incentives can amplify a misaligned proxy rather than repair it.

\section{L2 Additional Diagnostic: Flexible Capacity as a Coordination Test}\label{sec:app-l2}
The main-text L2 ablation turns on \kpi capacity steering and shows that adding flexible capacity can raise waiting. This appendix repeats the flex-pool sweep with capacity steering turned off (\(\xi=0\), \texttt{kpi\_steering\_mode=none}, so the flexible subpool is allocated by the static base-capacity shares rather than by last-period \kpi scores.

\begin{figure}[H]
  \centering
  \includegraphics[width=0.92\columnwidth]{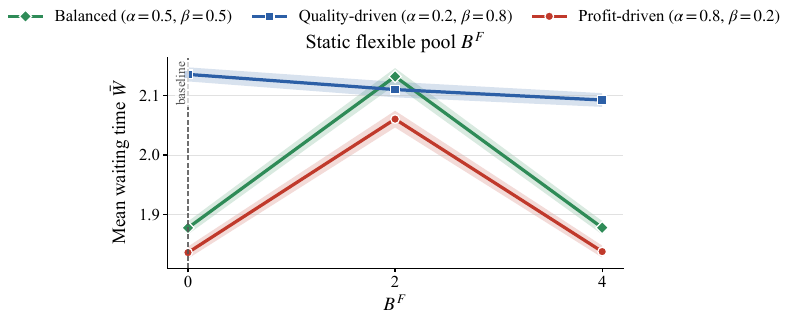}
  \caption{L2 steering-off flexible-pool diagnostic. Mean waiting (30 seeds, $T=200$) as a function of the flexible-pool size $B^{\mathrm{flex}}$, holding routing, team specialization, and total capacity fixed while disabling \kpi capacity steering. Three regimes are shown; the dashed line marks the L2 baseline.}
  \label{fig:l2staticflex}
\end{figure}

Figure~\ref{fig:l2staticflex} sweeps $B^{\mathrm{flex}} \in \{0,2,4\}$ under the three representative regimes of \S\ref{sec:l1}. With steering off, the adverse main-text slope disappears: mean waiting is essentially unchanged from $B^{\mathrm{flex}}=0$ to $4$ in the balanced regime ($1.88 \rightarrow 1.88$) and the profit-driven regime ($1.84 \rightarrow 1.84$), and falls modestly in the quality-driven regime ($2.14 \rightarrow 2.09$). The intermediate $B^{\mathrm{flex}}=2$ point rises in the balanced and profit-driven regimes, but we interpret this as a finite-team integer-allocation effect: total capacity is held fixed, so increasing $B^{\mathrm{flex}}$  capacity between dedicated and flexible slots, and a small flexible subpool can temporarily worsen queue--team mismatch before a larger subpool covers more teams. Therefore, flexible capacity by itself does not reproduce the adverse waiting increase seen in the main L2 panel; the increase comes from the \kpi-steering rule that consumes the flexible subpool.

When capacity steering is active, the flex pool is reallocated toward teams with high \kpi-steering scores, which in these rollouts are teams already adjusting triage and effort to chase bonus share. When steering is disabled, the flexible subpool is no longer systematically allocated to those teams, so it stops reinforcing the bonus-driven behavior that raised waiting in the main L2 ablation. The result shows that flexible capacity is not automatically beneficial: its effect depends on the allocation rule. In this configuration, $B^{\mathrm{flex}}$ helps only when the rule assigning it addresses queue imbalance rather than amplifying the coordination problem created by the \kpi tournament.

\section{L1/L2 Experimental Setup}\label{sec:app-l1-l2-setup}
Table~\ref{tab:l1-l2-setup} centralizes the run settings used by the L1 phase diagram and the L2 administrative-lever ablations. These settings are reported here for reproducibility; the main text uses only the mechanism-level findings.

\begin{table*}[t]
\centering
\scriptsize
\setlength{\tabcolsep}{3pt}
\renewcommand{\arraystretch}{1.14}
\begin{tabularx}{\textwidth}{@{}p{0.15\textwidth}p{0.25\textwidth}p{0.25\textwidth}X@{}}
\toprule
\textbf{Layer} &
\textbf{Scope} &
\textbf{Values} &
\textbf{Notes} \\
\midrule
L1/L2 &
Shared rollout protocol &
Horizon $T=200$; seeds $\{1,\ldots,30\}$; reported values are seed means unless a confidence interval is shown. &
Both layers use the native hospital administrator and provider-response rules of \S\ref{sec:env}; no \alphaevolve policy is injected. \\
L1 &
Incentive phase diagram &
$\alpha,\beta \in \{0.0,0.1,\ldots,1.0\}$, giving an $11\times 11$ grid. &
The sweep varies only provider financial sensitivity $\alpha$ and quality sensitivity $\beta$; all other administrative settings stay fixed. \\
L1 &
Default administrative and environment settings &
$B^{\mathrm{tot}}=10$, $B^{\mathrm{flex}}=0$, $B^{\mathrm{pool}}=5.0$, $\kappa=2.0$, $(w_H,w_W,w_{\mathrm{rej}},w_C)=(1.0,0.5,0.5,0.3)$, $\xi=0$. &
Strategic routing is disabled; \texttt{kpi\_steering\_mode=none}; probabilistic up-coding, \kpi-gaming tilt, and \kpi-weighted effort feedback are enabled. \\
L2 &
Representative regimes &
Profit-driven $(\alpha,\beta)=(0.8,0.2)$; quality-driven $(0.2,0.8)$; balanced $(0.5,0.5)$. &
Each L2 axis is swept one at a time around these three L1 anchor regimes. \\
L2 &
Main one-at-a-time lever sweeps &
$q\in\{0,0.05,0.10,0.20,0.35,0.50\}$; $B^{\mathrm{pool}}\in\{0,2.5,5,7.5,10,15\}$; $\kappa\in\{0,1,2,3,4,5\}$; $B^{\mathrm{tot}}\in\{6,8,10,12,14,16\}$; $w_H/w_C\in\{0.5,1,2,3.333,4,5\}$; $B^{\mathrm{flex}}\in\{0,2,4\}$. &
The audit sweep sets \texttt{audit\_base\_prob}$=q$ and \texttt{audit\_slope}$=0$; the $w_H/w_C$ sweep holds $w_H+w_C$ fixed at the L1 baseline. \\
L2 &
Main baseline and steering settings &
$B^{\mathrm{tot}}=10$, $B^{\mathrm{flex}}=0$, $B^{\mathrm{pool}}=5.0$, $\kappa=2.0$, $(w_H,w_W,w_{\mathrm{rej}},w_C)=(1.0,0.5,0.5,0.3)$, audit baseline $q=0.10$, $\xi=1.0$. &
Strategic routing is disabled; \texttt{kpi\_steering\_mode=none}; probabilistic up-coding, \kpi-targeting tilt, and \kpi-weighted effort feedback are enabled. \\
L2 &
Steering-off flexible-capacity diagnostic &
$B^{\mathrm{flex}}\in\{0,2,4\}$ under the same three regimes, with $B^{\mathrm{tot}}=10$, $\xi=0$, and \texttt{kpi\_steering\_mode=none}. &
This diagnostic isolates whether the adverse waiting slope comes from flexible capacity itself or from the \kpi-based allocation rule. \\
\bottomrule
\end{tabularx}
\caption{L1/L2 experimental setup and hyperparameters. $B^{\mathrm{tot}}$ is total capacity, $B^{\mathrm{flex}}$ is the flexible-capacity pool, and $B^{\mathrm{pool}}$ is the \kpi bonus pool.}
\label{tab:l1-l2-setup}
\end{table*}

\section{External Stylized-Fact Validation}\label{sec:app-external-validation}
Table~\ref{tab:external-validation-full} summarizes the external stylized facts used to validate \medisim{}'s provider-response dynamics. The table is not intended as a quantitative calibration exercise. It checks whether simulator interventions move the same provider-response channels in the same qualitative direction as the healthcare incentive and operations literature predicts.

\begin{table*}[t]
\centering
\scriptsize
\setlength{\tabcolsep}{3pt}
\renewcommand{\arraystretch}{1.12}
\begin{tabularx}{\textwidth}{@{}p{0.16\textwidth}p{0.13\textwidth}p{0.14\textwidth}p{0.16\textwidth}p{0.21\textwidth}X@{}}
\toprule
\textbf{External stylized fact} &
\textbf{Anchor} &
\textbf{Simulation intervention} &
\textbf{Expected qualitative signature} &
\textbf{\medisim observed signature} &
\textbf{Match strength and caveat} \\
\midrule
Diagnosis-based payment creates coding rent &
\citet{dafny2005hospitals}; \citet{kronick2014measuring} &
Increase profit sensitivity or payment spread; high-$\alpha$/low-$\beta$ region &
Reported complexity or up-coding rises; funds rise; true clinical need need not rise equally &
Profit-driven representative regime: up-coding $=0.226$; funds $=5451.7$ &
Strong directional match. We claim qualitative coding-rent reproduction, not real-system magnitude calibration. \\
\addlinespace[1pt]
Risk-adjusted reimbursement raises reported risk or coded complexity &
\citet{kronick2014measuring}; \citet{geruso2020upcoding} &
Strengthen reimbursement value of the coded group while the coding wedge is active &
Coded risk rises relative to true patient complexity &
High-profit cells show elevated up-coding; the main tables report the up-coding rate rather than coded-minus-true \cmi decomposition &
Partial match. Stronger if future reports add explicit coded-\cmi inflation. \\
\addlinespace[1pt]
Profit or ownership-like incentives shift case mix &
\citet{silverman2004medicare}; \citet{ellis1998creaming,ma1994health} &
Move toward high $\alpha$ at low $\beta$; compare profit-driven cells with quality and balanced cells &
More profitable or easier patients are favored; high-cost/high-\cmi patients face worse access &
Profit-driven representative regime: high-\cmi rejection gap $=0.182$ alongside elevated coding and funds &
Strong directional match. In L1/L2, active hospital routing is disabled, so the selection signature comes from provider triage. \\
\addlinespace[1pt]
Audit suppresses coding but can shift pressure elsewhere &
\citet{kuhn2008upcoding} &
Increase audit probability $q$ from 0 to 0.5 &
Up-coding falls; selection or delay pressure may rise when the billing channel closes &
Profit-driven up-coding $0.851\rightarrow0.003$; balanced up-coding $0.636\rightarrow0.001$; balanced cherry-picking $0.100\rightarrow0.233$ &
Strong match. The result validates channel substitution, not a calibrated audit schedule. \\
\addlinespace[1pt]
Measured performance targets induce gaming &
\citet{bevan2006measured,propper2010incentives}; \citet{campbell1979assessing,manheim2018goodhart} &
Increase \kpi salience through balanced-interior incentives, bonus pressure, or \kpi steering &
Measured score becomes behaviorally salient while true-health alignment can weaken &
Balanced interior: lagging teams tilt accepted case mix toward easier cases (tilt $=0.341$); $\mathrm{corr}(\mathrm{KPI},\mathrm{health})=-0.659$ &
Strong match. The simulator does not reproduce a specific NHS target; it reproduces proxy-objective decoupling. \\
\addlinespace[1pt]
Stronger pay-for-performance can worsen true alignment when the proxy is misaligned &
\citet{holmstrom1991multitask,baker1992incentive,eijkenaar2013p4p,vanherck2010p4p} &
Increase the bonus pool $B^{\mathrm{pool}}$ &
The measured proxy becomes more consequential; \kpi--health alignment may worsen &
Endpoint comparison: balanced $\mathrm{corr}(\mathrm{KPI},\mathrm{health})$ is $-0.447$ at $B^{\mathrm{pool}}=0$ and $-0.839$ at $B^{\mathrm{pool}}=15$; the smallest positive pool gives a small non-monotone uptick &
Strong match. The caveat is conditional: stronger incentives are harmful here because the measured proxy is misaligned. \\
\addlinespace[1pt]
Quality-oriented incentives suppress coding/selection but raise effort and budget pressure &
\citet{ma1994health,ellis1998creaming,eggleston2005multitasking} &
Low $\alpha$, high $\beta$ regime &
Lower up-coding and less cream-skimming; higher effort and weaker solvency &
Quality-driven regime: up-coding $=0.027$; rejection gap $=-0.051$; effort $=2.435$; funds $=15.2$ &
Strong match. This is a multitasking trade-off in the simulator's simplified clinical-production function. \\
\addlinespace[1pt]
Total capacity expansion reduces waiting under queueing pressure &
\citet{green2002beds,green2006queueing} &
Increase total capacity from 6 to 16 &
Mean waiting should fall across regimes &
Mean waiting falls in all three L2 regimes: profit $2.313\rightarrow1.735$, quality $2.325\rightarrow1.896$, balanced $2.303\rightarrow1.720$ &
Strong operations sanity check. This row validates the queueing layer rather than a strategic distortion channel. \\
\addlinespace[1pt]
Flexible capacity helps only under an appropriate allocation rule &
\citet{bekker2017flexible} and hospital-flow allocation logic &
Compare \kpi-steered flexible pool with steering-off flexible pool &
Flex capacity may fail if allocated by the wrong proxy; the adverse effect should disappear when steering no longer follows the proxy &
With \kpi steering, balanced waiting rises $1.88\rightarrow2.23$ as $B^{\mathrm{flex}}$ grows $0\rightarrow4$. With steering off, balanced and profit-driven waiting remain $1.88\rightarrow1.88$ and $1.84\rightarrow1.84$; quality falls $2.14\rightarrow2.09$ &
Strong special finding. Total capacity is fixed here; the experiment tests the allocation rule for the flexible subpool. \\
\bottomrule
\end{tabularx}
\caption{Full external stylized-fact validation map. Rows report qualitative matches between external healthcare incentive/operations facts and \medisim signatures. Match strength refers to directional reproduction under fixed simulator rules, not quantitative calibration to any real hospital system.}
\label{tab:external-validation-full}
\end{table*}

\section{Strategic Policy-as-Code: Search Diagnostics}\label{sec:app-alphaevolve}
This part provides four pieces of structural evidence behind the L3 results of \S\ref{sec:l3}: validation search curves, a warm-start ablation that pinpoints the role of the curated library, method comparisons under each social objective that contextualize the headline numbers in Table~\ref{tab:l3}, and a $K=300$ diagnostic that characterizes the shape of the search trajectory.

The main L3 runs use ChatGPT-5.4 as the code-mutation model with $K=200$ iterations, 3 islands, 30 individuals per island, LLM temperature 0.4, migration size 2, evolution seeds $\{101,202,303\}$, validation seeds $\{404,505,606,707,808\}$, and held-out test seeds $\{909,1001,1103,1207,1301\}$. The warm-start library and its diversity ablations are described in \S\ref{sec:app-alphaevolve-warmstart}.

\begin{figure*}[!t]
  \centering
  \includegraphics[width=0.96\textwidth]{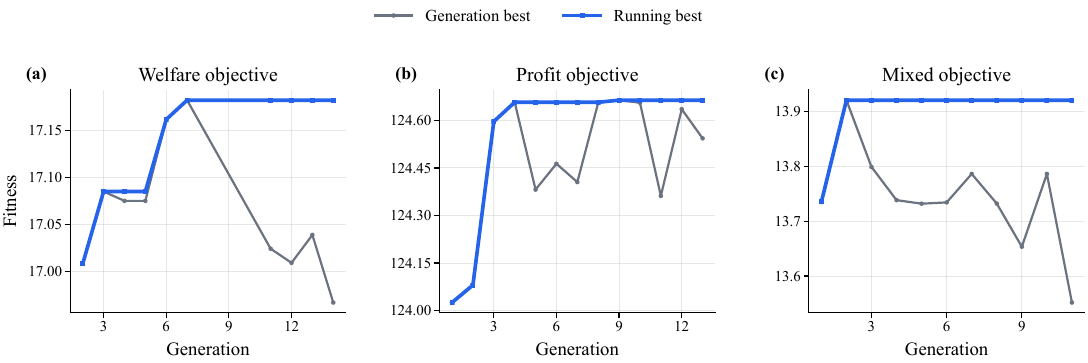}
  \caption{L3 \alphaevolve validation search curves. Each panel shows generation-best and running-best fitness under a different evaluator with K=200. These curves are diagnostic: the main-text L3 claim rests on held-out rollout profiles in Table~\ref{tab:l3}, while the curves show how each objective shapes the search trajectory.}
  \label{fig:l3curves}
\end{figure*}

Figure~\ref{fig:l3curves} records the validation dynamics for the $K=200$ runs. Welfare and profit objectives show short early improvements followed by plateaus, while the mixed objective reaches its best validation candidate quickly and then mostly explores lower-fitness variants. We use these curves as search-process diagnostics; the policy comparison itself is based on the held-out rollout profiles in Table~\ref{tab:l3}.

\subsection{Warm-start library and the diversity ablation}\label{sec:app-alphaevolve-warmstart}
The L3 warm-start library contains nine policies: a \emph{neutral} template (constants equal to balanced defaults), a \emph{fixed} heuristic that mirrors the L1 baseline, a \emph{greedy-profit} policy with high $\alpha$ and aggressive coding, a \emph{greedy-quality} policy with high $\beta$ and conservative coding, three access-oriented variants (high-$\alpha$ acceptance, high-capacity, high-flex), and two coding-aggression variants (aggressive and conservative). Library diversity is the only configuration parameter that meaningfully changes the mixed-objective outcome at $K=200$.

The ablation is reported as three nested libraries.
With the \emph{neutral-only} library, selection on validation fitness returns the seed itself at $13.545$ and the best evolved candidate reaches only $13.351$; search cannot improve over its starting point.
After removing the profit-side seeds, which the library still contains the welfare-leaning policies but no aggressive-coding warm start, the process produces the best evolved candidate at $13.401$.
The \emph{full} nine-policy library produces a selected aggressive-coding warm start at $13.607$ and then refines to $13.876$ on held-out seeds, the value reported in Table~\ref{tab:l3}.
The reading we take from this nested set is clear: the gain from search over the best warm start ($+0.27$ fitness) is comparable to the gain from library expansion alone ($+0.21$), and search produces no improvement above the seed when the seed is not diverse. We therefore present \alphaevolve over \medisim as a feasibility demonstration of program search over the policy class of \S\ref{sec:mac} and do not claim that current search procedures can rediscover the mixed family from scratch.

\subsection{Method comparison under each social objective}\label{sec:app-alphaevolve-method}
\begin{figure*}[!t]
  \centering
  \includegraphics[width=0.96\textwidth]{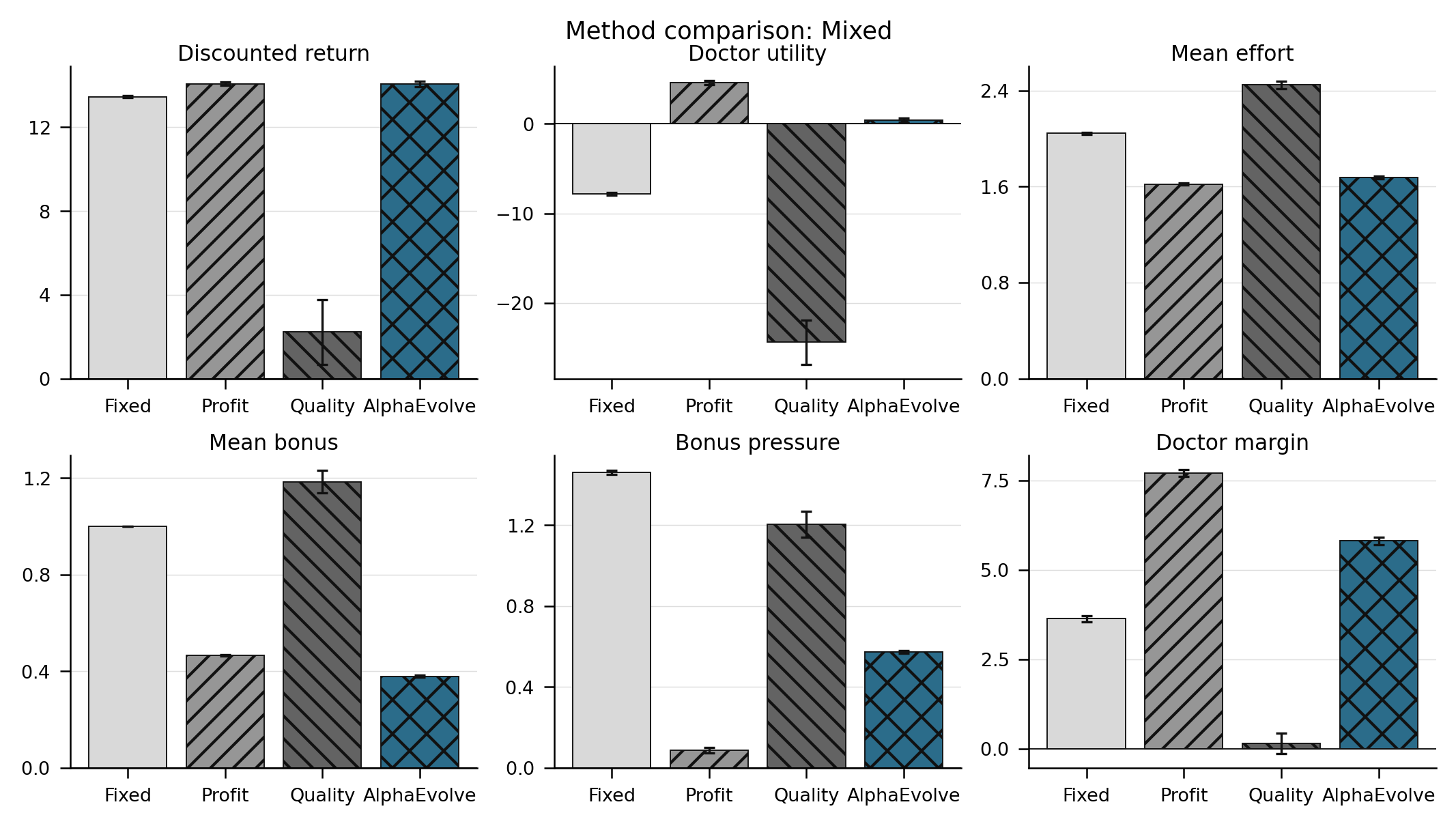}
  \caption{$K=200$ mixed-objective four-method comparison. Bars are means over the five held-out test seeds; whiskers are 95\% bootstrap intervals. The \alphaevolve column is the central qualitative finding: it achieves profit-comparable discounted return while reducing mean bonus and bonus pressure by about 60\% relative to \textsc{Fixed}, with doctor margin recovered to within 25\% of the \textsc{Profit} baseline.}
  \label{fig:app-alphaevolve-mixed-k200}
\end{figure*}

Figure~\ref{fig:app-alphaevolve-mixed-k200} is the main supporting figure for the mixed-objective L3 result. The discounted-return panel shows \alphaevolve essentially tied with the profit baseline; the doctor-utility, mean-effort, and doctor-margin panels show that this tie is achieved with a substantially different operational profile: lower mean effort, higher doctor utility, and higher doctor margin than \textsc{Fixed}. The two bonus-related panels are the most informative. Mean bonus and bonus pressure both fall by about 60\% relative to the \textsc{Fixed} baseline, even though \textsc{Fixed} and \alphaevolve produce comparable discounted returns. The searched mixed policy therefore retains the macroeconomic return while structurally lowering the local bonus pressure $B^{\mathrm{pool}}\kappa s_j(1-s_j)$ that drives the \kpi-targeting residual identified in \S\ref{sec:app-l1-kpi}.

\begin{figure*}[!t]
  \centering
  \includegraphics[width=0.96\textwidth]{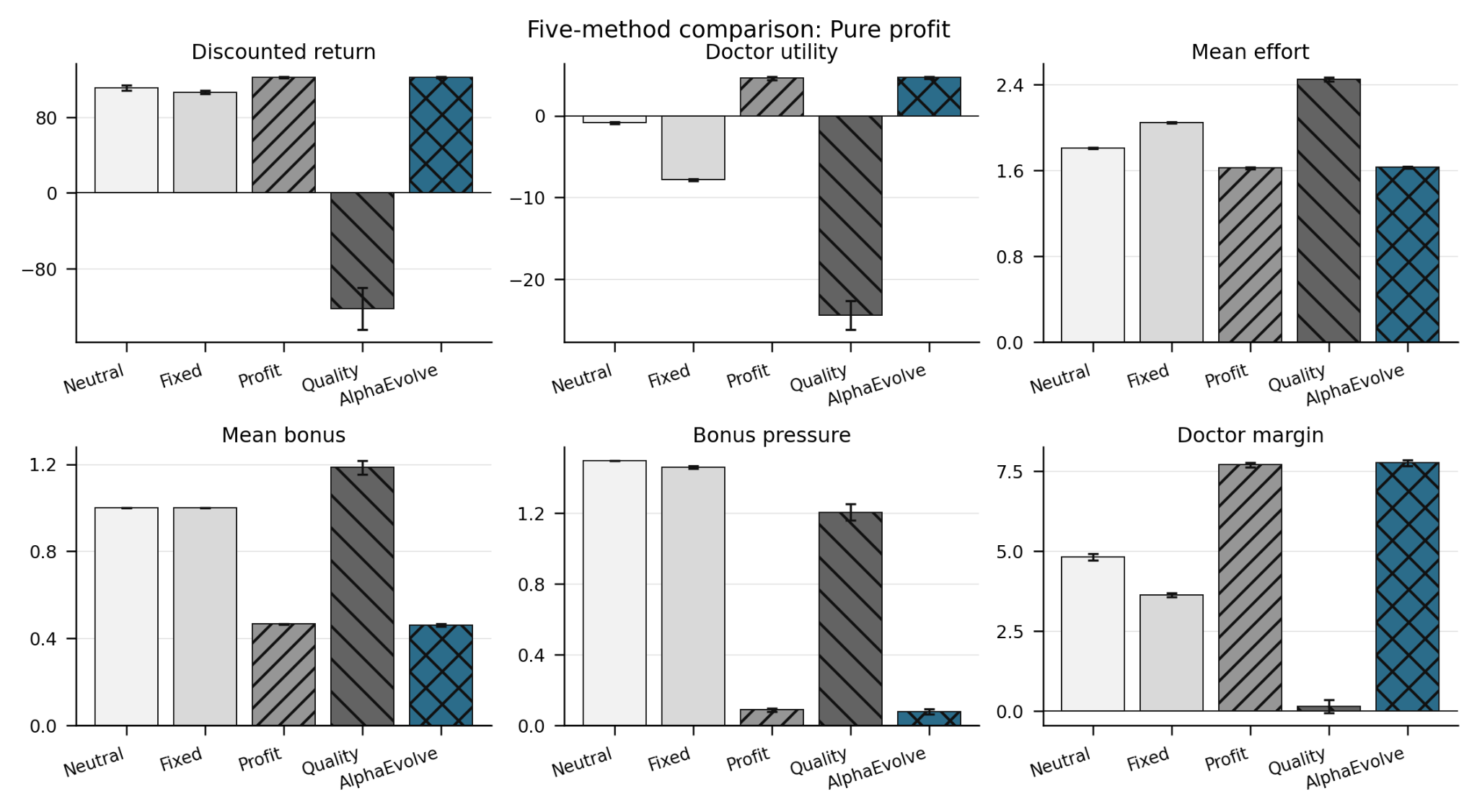}
  \caption{$K=200$ pure-profit five-method comparison. Bars are means over the five held-out test seeds. Adding the \textsc{Neutral} baseline reveals that \textsc{Neutral} achieves a lower but still nontrivial return despite substantially higher mean effort and bonus pressure; \alphaevolve makes a small profit-family refinement visible in return, doctor utility, and doctor margin.}
  \label{fig:app-alphaevolve-profit-five}
\end{figure*}

\begin{figure*}[!t]
  \centering
  \includegraphics[width=0.96\textwidth]{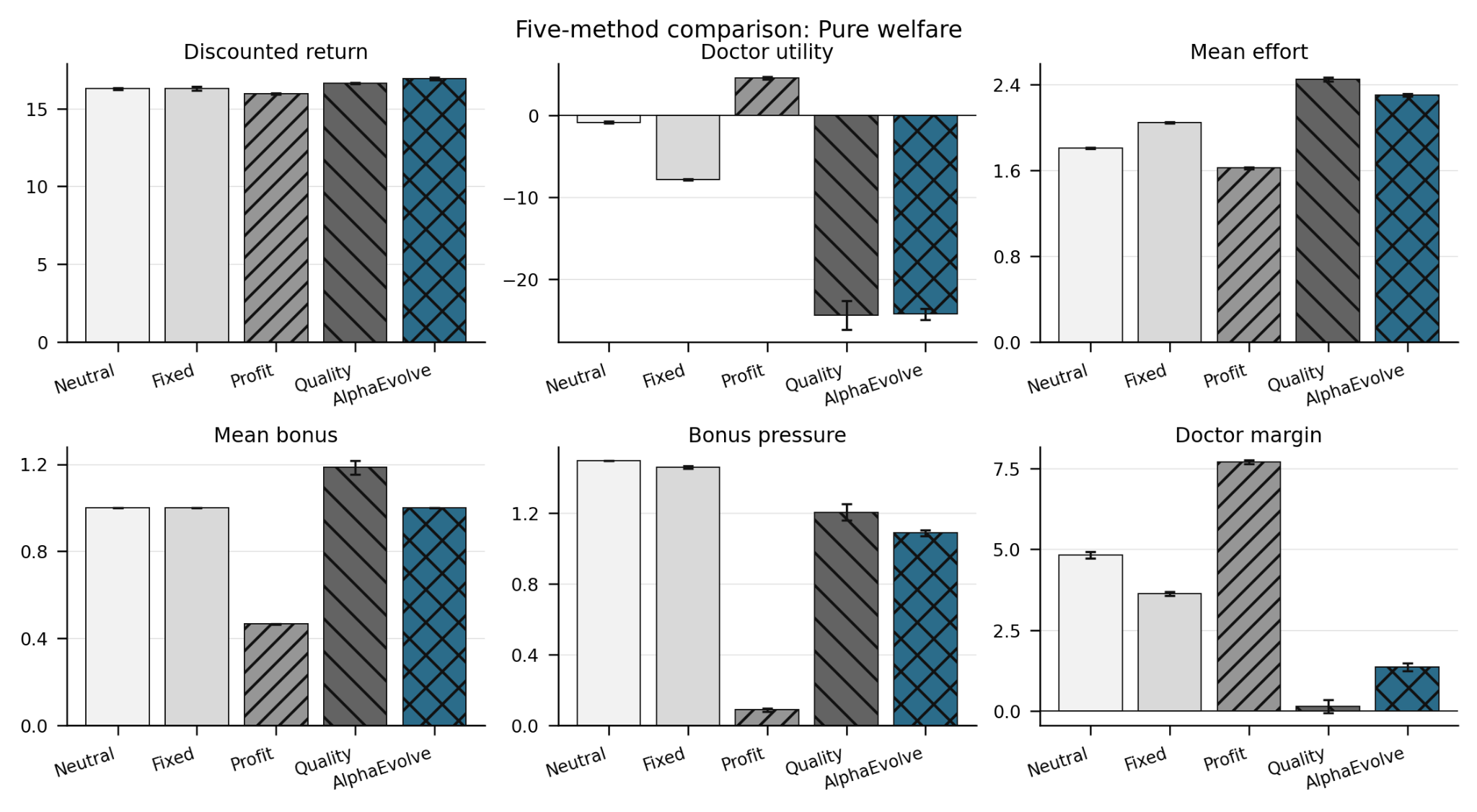}
  \caption{$K=200$ pure-welfare five-method comparison. Bars are means over the five held-out test seeds. The \alphaevolve refinement of the welfare family lifts discounted return above all four baselines while keeping mean effort below \textsc{Quality} and lifting doctor margin from near zero to $\sim\!1.4$, indicating that the search has trimmed the gold-plating slack identified in \S\ref{sec:app-l1-clinical}.}
  \label{fig:app-alphaevolve-welfare-five}
\end{figure*}

The single-objective comparisons in Figures~\ref{fig:app-alphaevolve-profit-five} and~\ref{fig:app-alphaevolve-welfare-five} sharpen the interpretive picture in two ways. Under the pure-profit objective, the \textsc{Profit} warm start is already close to a local optimum on $\mathrm{Fitness}$; \alphaevolve captures only a small additional margin, visible as minor gains in discounted return, doctor utility, and doctor margin, while pushing the coding-heavy risk profile slightly further. Adding the \textsc{Neutral} baseline shows that the return gap between \textsc{Neutral} and \textsc{Profit} is moderate, but the operational gap is much larger: \textsc{Neutral} uses substantially more effort, bonus, and bonus pressure while still remaining below the profit-oriented return. Under the pure-welfare objective, \alphaevolve also improves the operational profile relative to the Quality baseline. The doctor-margin panel rises from near zero to about $1.4$, indicating that the searched welfare policy reduces the excessive effort cost of the quality-driven baseline. At the same time, it preserves the access improvements rewarded by the welfare objective, rather than gaining margin by rejecting or delaying patients.

\subsection{\texorpdfstring{$K=300$}{K=300} comparison panels}\label{sec:app-alphaevolve-k300comp}
\begin{figure*}[!t]
  \centering
  \includegraphics[width=0.96\textwidth]{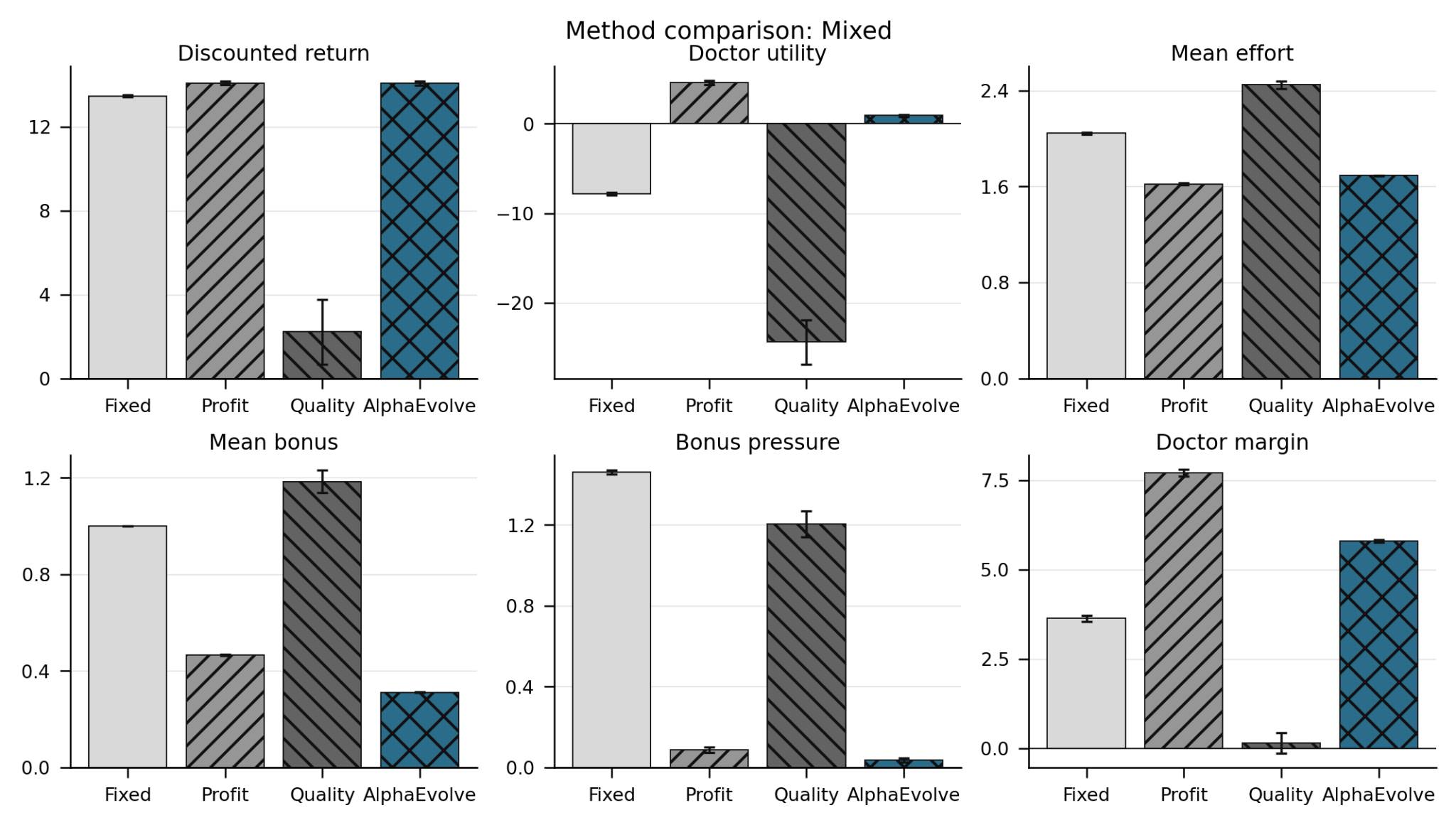}
  \caption{$K=300$ mixed-objective four-method comparison, the longer-budget counterpart of Figure~\ref{fig:app-alphaevolve-mixed-k200}. Bars are means over the five held-out test seeds. The qualitative picture is unchanged from $K=200$: \alphaevolve achieves discounted return on par with \textsc{Profit} while leaving mean bonus and bonus pressure suppressed, consistent with the trajectory in Figure~\ref{fig:app-alphaevolve-mixed-steps}.}
  \label{fig:app-alphaevolve-mixed-comp}
\end{figure*}

\begin{figure*}[!t]
  \centering
  \includegraphics[width=0.96\textwidth]{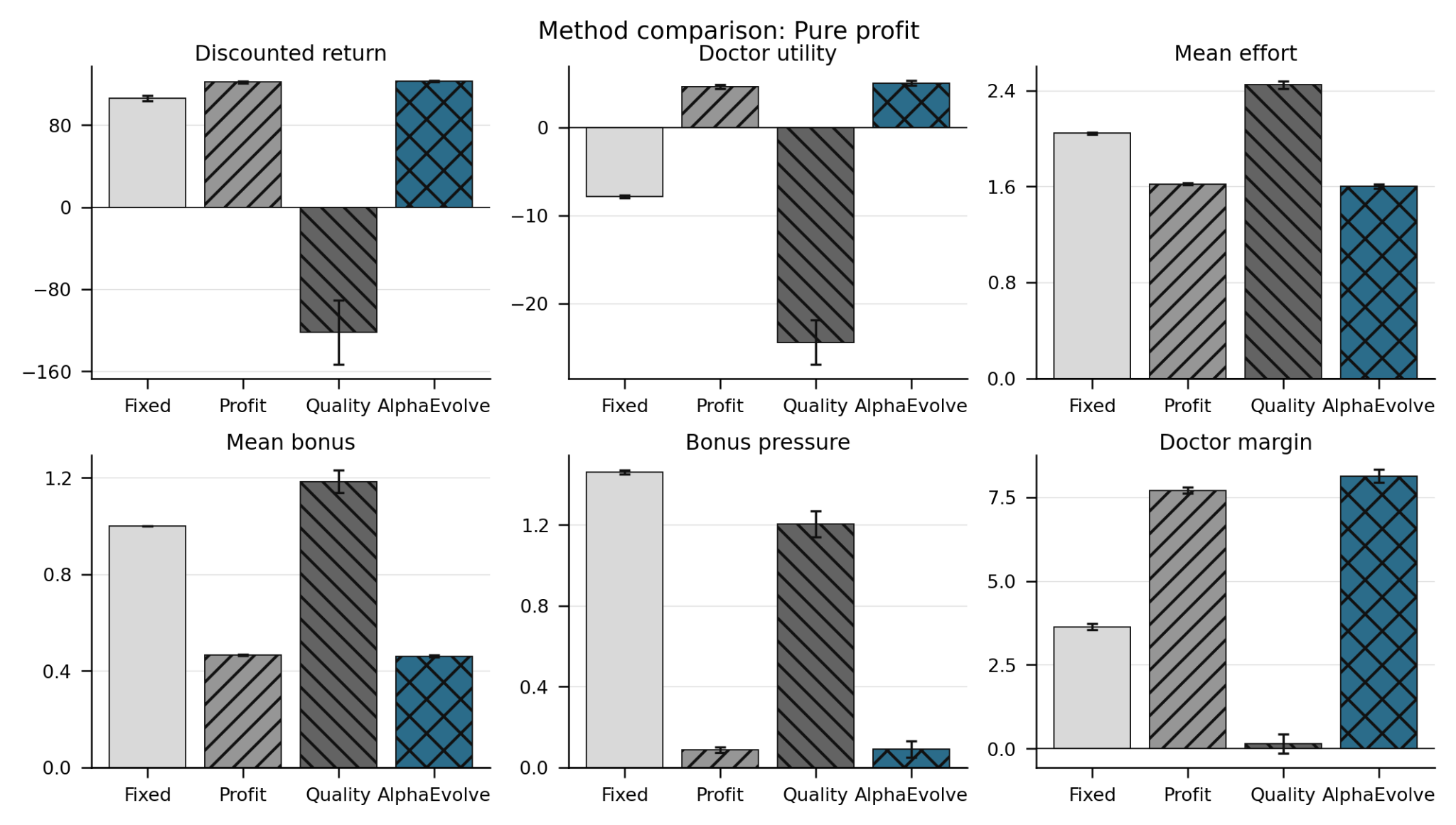}
  \caption{$K=300$ pure-profit four-method comparison. Bars are means over the five held-out test seeds. The longer budget gives \alphaevolve a small gain over \textsc{Profit} in discounted return, doctor utility, and doctor margin, while preserving the same profit-oriented risk profile.}
  \label{fig:app-alphaevolve-profit-comp}
\end{figure*}

\begin{figure*}[!t]
  \centering
  \includegraphics[width=0.96\textwidth]{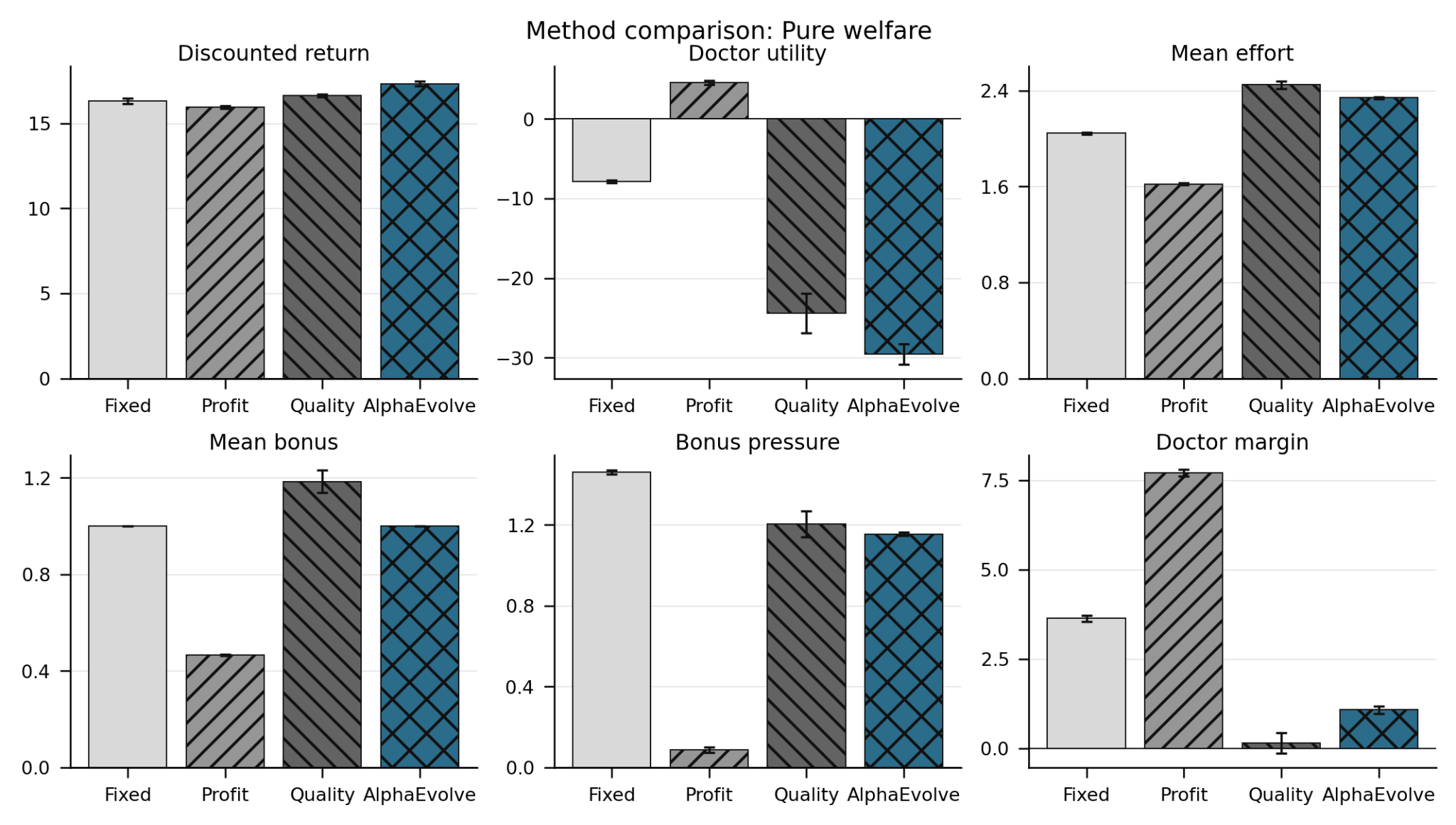}
  \caption{$K=300$ pure-welfare four-method comparison. Bars are means over the five held-out test seeds. The welfare \alphaevolve refinement extends the discounted-return lift seen at $K=200$ and continues to suppress \textsc{Quality}-style gold-plating on the effort and bonus panels.}
  \label{fig:app-alphaevolve-welfare-comp}
\end{figure*}

The four-method comparisons in Figures~\ref{fig:app-alphaevolve-mixed-comp}--\ref{fig:app-alphaevolve-welfare-comp} reproduce the same qualitative policy families at the longer search budget. The \textsc{Fixed}, \textsc{Profit}, and \textsc{Quality} bars are unchanged because they are evaluations of fixed warm starts. The \alphaevolve bars move in objective-specific directions: the pure-profit and pure-welfare runs push further along their respective objectives, while the mixed run improves waiting and high-complexity deferral but does not dominate the $K=200$ mixed policy on held-out fitness or aggregate violations. We therefore treat $K=300$ as a diagnostic budget check rather than as a replacement for the main $K=200$ result.

\subsection{\texorpdfstring{$K=300$}{K=300} diagnostic: trajectory shape}\label{sec:app-alphaevolve-k300}
\begin{figure*}[!t]
  \centering
  \includegraphics[width=0.62\textwidth]{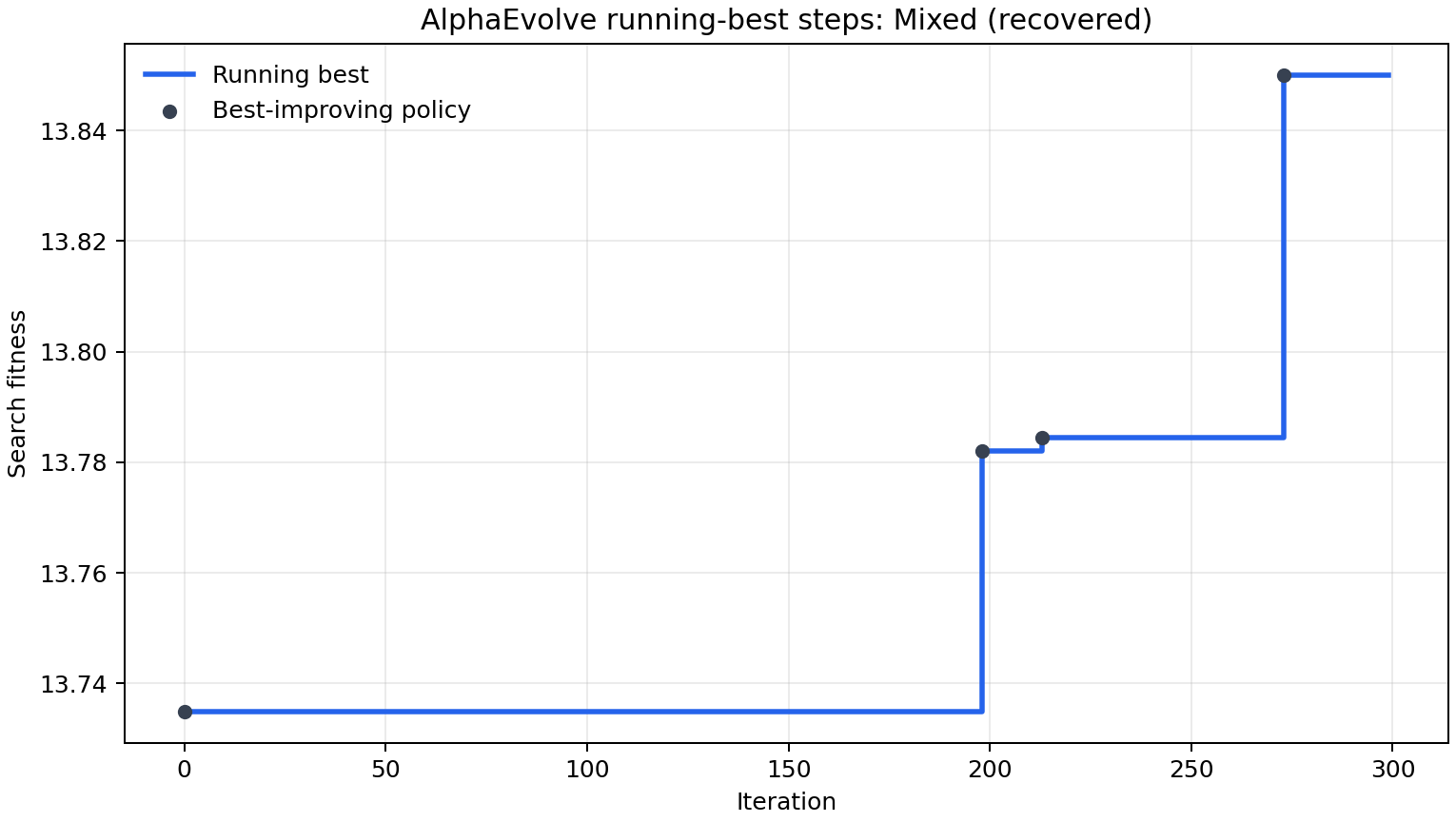}
  \caption{$K=300$ mixed-objective running-best trace. Each step is a search iteration; the first marker is the neutral seed and later markers indicate iterations at which \alphaevolve discovered a policy that improved the incumbent search fitness. The trajectory is piecewise constant with three update events at iterations $198$, $213$, and $273$; Table~\ref{tab:k300-policy-trace} records the code-level edits associated with each event.}
  \label{fig:app-alphaevolve-mixed-steps}
\end{figure*}

Extending the search budget from $K=200$ to $K=300$ does not change the held-out conclusion: the $K=300$ mixed run lowers high-complexity deferral but does not beat the $K=200$ policy on held-out fitness or overall violations, and we therefore keep $K=200$ as the main mixed result. The $K=300$ trace is nonetheless diagnostic about the \emph{shape} of program-space search over $\Pi_A$. Figure~\ref{fig:app-alphaevolve-mixed-steps} shows that improvement is concentrated in three discrete events at iterations $198$, $213$, and $273$, separated by long intervals of zero improvement. The fitness increments are small ($+0.047$, $+0.002$, and $+0.066$), and only two of the three events touch the administrative side of the policy bundle. This piecewise-constant trajectory shape is a feature of the regulated DSL of \S\ref{sec:mac}: the program space contains many syntactically valid local perturbations whose simulated rollouts produce indistinguishable evaluator returns, and meaningful improvements arrive only at the rare iterations where a coordinated multi-field edit moves the mechanism across one of the regime boundaries identified in \S\ref{sec:app-l1}.

\begin{table}[H]
  \caption{$K=300$ mixed running-best trace. Update codes summarize the policy fields changed relative to the previous running-best step. ``Gap'' is the strategic-delay gap.}
  \label{tab:k300-policy-trace}
  \centering
  \scriptsize
  \resizebox{\columnwidth}{!}{%
  \begin{tabular}{rrlrrrr}
    \toprule
    Step & Iter. & Update & Fitness & Wait & Reject & Gap \\
    \midrule
    1 & 0 & neutral & 13.735 & 1.797 & 0.022 & 0.328 \\
    2 & 198 & admin + doctor & 13.782 & 1.628 & 0.050 & 0.233 \\
    3 & 213 & doctor only & 13.784 & 1.606 & 0.050 & 0.236 \\
    4 & 273 & admin + doctor & 13.850 & 1.602 & 0.049 & 0.238 \\
    \bottomrule
  \end{tabular}
  }
\end{table}

Table~\ref{tab:k300-policy-trace} decomposes these update events at the metric level. Each improvement event corresponds to a clear trade: rejection rises from $0.022$ to $\sim 0.05$ as the policy stops absorbing patients it cannot treat profitably, the strategic-delay gap falls from $0.328$ to $\sim 0.24$ as the rejection channel takes over the cost-shedding role the delay channel had carried, and mean waiting falls in lockstep. The qualitative pattern is the same channel substitution documented in \S\ref{sec:app-l1-access}, now observed inside a single search trajectory. The complete diagnostic policy sketch is given in Listing~\ref{lst:k300-mixed-sketch} below.

\subsection{Discovered policy structures}\label{sec:app-alphaevolve-listings}
The first three listings below summarize the final code structures produced by the $K=200$ search under each social objective. Each listing reproduces the final field-level policy code; auxiliary dictionary syntax is omitted for readability and the full executable policies are saved by the experiment pipeline. The structures should be read against the warm-start ablation of \S\ref{sec:app-alphaevolve-warmstart}: search refines, but does not invent, the families inherited from the library.

\paragraph{Welfare family (Listing~\ref{lst:pure-welfare-sketch}).}
The welfare family pushes $\alpha$ down to $0.20$ and $\beta$ to $1.00$, then adds state-conditional capacity expansion ($+1$ if any of queue, waiting, or utilization stress is detected) and a flex pool that grows under the same conditions. The hospital \kpi weight vector tilts toward health, waiting, and rejection ($w_H=1.80$, $w_W=0.90$, $w_{\mathrm{rej}}=1.70$) while suppressing the cost weight to $w_C=0.05$. The coding-side expression is dominated by the ethics term ($-2.50$), which keeps up-coding at zero across all rollouts.

\begin{lstlisting}[caption={Pure-welfare policy sketch.},label={lst:pure-welfare-sketch}]
alpha = 0.20
beta = 1.00
total_capacity = max(9, min(13, current_total_capacity
    + I((queue_total > 6.0) or (wbar > 1.4) or (utilization > 0.88))
    - I((queue_total < 2.0) and (utilization < 0.50))))
bonus_pool = 5.0
flex_pool = min(max(1, 3 + I((wbar > 1.2) or (queue_total > 5.0)
    or (queue_max > 2.0)) + I(utilization > 0.85)),
    max(1, current_total_capacity))
xi = 0.0
wH = 1.80; wW = 0.90; wrej = 1.70; wC = 0.05; kappa = 1.7
effort = base_effort + effort_quality_pressure
triage_accept = (clinical_triage_score + margin_pressure + quality_pressure
    + bed_availability_signal - triage_fatigue_pressure + kpi_gaming_pressure
    + 0.15 * urgency + 0.05 * wait) >= accept_threshold
triage_defer = (not ((clinical_triage_score + margin_pressure + quality_pressure
    + bed_availability_signal - triage_fatigue_pressure + kpi_gaming_pressure
    + 0.15 * urgency + 0.05 * wait) >= accept_threshold)) and
    ((clinical_triage_score + margin_pressure + quality_pressure
    + bed_availability_signal - triage_fatigue_pressure
    + 0.5 * kpi_gaming_pressure + 0.10 * urgency) + defer_pressure
    >= defer_threshold)
request_bed = bed_available and ((clinical_bed_score + 0.20 * quality_pressure
    + bed_quality_pressure - bed_cost_pressure + bed_availability_signal
    - bed_fatigue_pressure + 0.05 * urgency) >= bed_threshold)
candidate_score = 0.20 * upcode_pressure - 1.50 * audit_penalty
    - 2.50 * ethics_pressure - 0.45 * coding_gap
\end{lstlisting}

\paragraph{Profit family (Listing~\ref{lst:pure-profit-sketch}).}
The profit family inverts every one of these moves. The hospital administrator commits to $\alpha \approx 0.95$ except when waiting or rejection signals a near-failure mode (in which case $\alpha$ falls and $\beta$ rises by $0.15$--$0.20$ as a safety reflex), keeps \kpi steering active at $\xi=1.5$, and lets a small flex pool expand under queue and rejection stress. The bonus pool scales linearly with last-period profit, and the hospital \kpi weight on cost ($w_C=1.70$) dominates all other weights. The coding-side expression doubles its weight on up-coding pressure and gain and halves its ethics-pressure coefficient: the searched profit policy actively manages coding aggressiveness as a continuous variable.

\begin{lstlisting}[caption={Pure-profit policy sketch.},label={lst:pure-profit-sketch}]
alpha = clip(0.95 - 0.15 * I((wbar > 4.0) or (rbar > 0.15)), 0.0, 1.0)
beta = clip(0.05 + 0.20 * I((wbar > 4.0) or (rbar > 0.15)), 0.0, 1.0)
total_capacity = max(10, min(12, current_total_capacity
    + I((((wbar > 5.0) or (rbar > 0.18) or (queue_total > 24.0))
        and (utilization > 0.92)) and (funds > 70.0)
        and (profit_last > 0.0))
    - I((utilization < 0.62) and (wbar < 0.5)
        and (current_total_capacity > 10))))
bonus_pool = clip(1.20 + 0.032 * max(0.0, profit_last)
    - 0.70 * I((profit_last < 0.0) or (rbar > 0.14)), 0.25, 4.1)
flex_pool = min(2, min(max(1, current_total_capacity),
    max(0, current_flex_pool
        + I(((queue_total > 9.0) or (queue_max > 4.0)) and (rbar > 0.05))
        - I((not (((queue_total > 9.0) or (queue_max > 4.0))
            and (rbar > 0.05))) and (current_flex_pool > 0)))))
xi = 1.5
wH = 0.15; wW = 0.05; wrej = 0.15; wC = 1.70; kappa = 3.5
effort = base_effort + effort_quality_pressure
triage_accept = (clinical_triage_score + margin_pressure + quality_pressure
    + bed_availability_signal - triage_fatigue_pressure + kpi_gaming_pressure
    + 0.22 * urgency + 0.06 * wait) >= accept_threshold
triage_defer = (not ((clinical_triage_score + margin_pressure + quality_pressure
    + bed_availability_signal - triage_fatigue_pressure + kpi_gaming_pressure)
    >= accept_threshold)) and ((clinical_triage_score + margin_pressure
    + quality_pressure + bed_availability_signal - triage_fatigue_pressure
    + kpi_gaming_pressure) + defer_pressure >= defer_threshold)
request_bed = bed_available and ((clinical_bed_score + bed_quality_pressure
    - bed_cost_pressure + bed_availability_signal - bed_fatigue_pressure)
    >= bed_threshold)
candidate_score = 0.22 * upcode_gain + 0.78 * upcode_pressure
    - 1.14 * audit_penalty - 1.16 * ethics_pressure - 0.26 * coding_gap
\end{lstlisting}

\paragraph{Mixed family (Listing~\ref{lst:k200-mixed-sketch}).}
The calibrated mixed family is the cleanest qualitative result of the L3 layer and the only policy in our search that simultaneously achieves three properties: profit-comparable discounted return, zero up-coding on held-out rollouts, and a halving of rejection relative to the profit baseline. Its structure is informative. The hospital administrator fixes $\alpha=0.5$, adapts $\beta$ upward only when access stress is detected, and leaves total capacity, the flex pool, and \kpi steering unchanged; the bonus pool grows with both profit and access stress but contracts sharply under insolvency; hospital \kpi weights are balanced with a primary tilt toward health and rejection. The candidate-coding score is the structural innovation: it carries a moderate positive weight on $\mathtt{upcode\_pressure}$ as a continuous lever, balanced against an explicit $-100$ indicator penalty that snaps to a regime that strictly forbids coding deviations above $0.20$. This indicator term is what drives up-coding to exactly zero on held-out seeds while leaving the gradient signal usable in early-training rollouts---a hand-engineered version of the same hard-threshold-plus-smooth-shaping idiom that has emerged in several published reward-design studies.

\begin{lstlisting}[caption={$K=200$ calibrated mixed policy sketch.},label={lst:k200-mixed-sketch}]
alpha = 0.5
beta = clip(0.35 + 0.20 * I((wbar > 4.0) or (rbar > 0.15)), 0.0, 1.0)
bonus_pool = clip(1.45 + 0.012 * max(0.0, profit_last)
    + 0.25 * I((wbar > 1.8) or (rbar > 0.06))
    - 0.55 * I(profit_last < 0.0), 0.75, 4.25)
total_capacity = current_total_capacity
flex_pool = current_flex_pool
xi = 0.0
wH = 1.0; wW = 0.20; wrej = 1.10; wC = 0.10; kappa = 2.0
effort = base_effort + effort_quality_pressure
triage_accept = (clinical_triage_score + margin_pressure + quality_pressure
    + bed_availability_signal - triage_fatigue_pressure
    + kpi_gaming_pressure) >= accept_threshold
triage_defer = (not ((clinical_triage_score + margin_pressure + quality_pressure
    + bed_availability_signal - triage_fatigue_pressure
    + kpi_gaming_pressure) >= accept_threshold)) and ((clinical_triage_score
    + margin_pressure + quality_pressure + bed_availability_signal
    - triage_fatigue_pressure + kpi_gaming_pressure) + defer_pressure
    >= defer_threshold)
request_bed = bed_available and ((clinical_bed_score + bed_quality_pressure
    - bed_cost_pressure + bed_availability_signal - bed_fatigue_pressure)
    >= bed_threshold)
candidate_score = 0.85 * upcode_pressure + 0.08 * upcode_gain
    - 1.00 * audit_penalty - 0.85 * ethics_pressure
    - 0.15 * coding_gap - 100.0 * I(coding_gap > 0.20)
\end{lstlisting}

\paragraph{$K=300$ diagnostic trace (Listing~\ref{lst:k300-mixed-sketch}).}
The $K=300$ trace produces a closely related but distinct mixed family. Listing~\ref{lst:k300-mixed-sketch} gives the neutral seed and then records only the code-level deltas at each running-best update in Table~\ref{tab:k300-policy-trace}; the complete executable policies are stored in the experiment artifacts. The main edits occur in two stages: iteration $198$ moves the administrative policy into an access-stress rule with active \kpi steering, and iterations $213$ and $273$ refine the bonus, rejection-weight, triage, and coding expressions. The $K=300$ family improves waiting and high-complexity deferral, but it does not dominate the $K=200$ family on held-out fitness or the aggregate violation score; we therefore keep $K=200$ as the main result, while using the trace to show that both budgets converge on the same coding-gap idiom.

\begin{lstlisting}[caption={$K=300$ mixed running-best trace deltas.},label={lst:k300-mixed-sketch},basicstyle=\ttfamily\tiny]
# Iteration 0: neutral seed.
alpha = 0.5; beta = 0.5
total_capacity = current_total_capacity; flex_pool = current_flex_pool
bonus_pool = 5.0; xi = 0.0
wH = 1.0; wW = 0.20; wrej = 1.0; wC = 0.10; kappa = 2.0
triage_accept = (clinical_triage_score + margin_pressure + quality_pressure
    + bed_availability_signal - triage_fatigue_pressure
    + kpi_gaming_pressure) >= accept_threshold
triage_defer = (not ((clinical_triage_score + margin_pressure + quality_pressure
    + bed_availability_signal - triage_fatigue_pressure
    + kpi_gaming_pressure) >= accept_threshold)) and ((clinical_triage_score
    + margin_pressure + quality_pressure + bed_availability_signal
    - triage_fatigue_pressure + kpi_gaming_pressure) + defer_pressure
    >= defer_threshold)
request_bed = bed_available and ((clinical_bed_score + bed_quality_pressure
    - bed_cost_pressure + bed_availability_signal - bed_fatigue_pressure)
    >= bed_threshold)
candidate_score = upcode_pressure - audit_penalty - ethics_pressure

# Iteration 198: delta relative to neutral seed.
alpha -> clip(0.65 - 0.08 * I((wbar > 4.0) or (rbar > 0.15)), 0.0, 1.0)
beta  -> clip(0.30 + 0.25 * I((wbar > 4.0) or (rbar > 0.15)), 0.0, 1.0)
total_capacity -> max(10, min(12, current_total_capacity
    + I((((wbar > 5.0) or (rbar > 0.18) or (queue_total > 24.0))
        and (utilization > 0.92)) and (funds > 70.0)
        and (profit_last > 0.0))
    - I((utilization < 0.62) and (wbar < 0.5)
        and (current_total_capacity > 10))))
bonus_pool -> clip(1.50 + 0.015 * max(0.0, profit_last)
    - 0.50 * I(profit_last < 0.0), 0.75, 4.25)
flex_pool -> min(2, min(max(1, current_total_capacity),
    max(0, current_flex_pool
        + I((queue_total > 10.0) and ((rbar > 0.06) or (wbar > 2.0)))
        - I((not ((queue_total > 10.0)
            and ((rbar > 0.06) or (wbar > 2.0))))
            and (current_flex_pool > 0)))))
xi -> 1.5
wH,wW,wrej,wC,kappa -> 0.85, 0.05, 1.05, 0.80, 2.2
triage_accept: add 0.14 * urgency + 0.04 * wait
triage_defer: add 0.14 * urgency + 0.04 * wait to accept gate;
              add 0.08 * urgency + 0.03 * wait to defer score
request_bed: add 0.08 * urgency + 0.04 * wait
candidate_score -> 0.85 * upcode_pressure + 0.08 * upcode_gain
    - 1.00 * audit_penalty - 0.85 * ethics_pressure
    - 0.15 * coding_gap - 100.0 * I(coding_gap > 0.20)

# Iteration 213: delta relative to iteration 198.
triage_accept: urgency coefficient 0.14 -> 0.16
triage_defer: accept-gate urgency 0.14 -> 0.16;
              defer-score urgency 0.08 -> 0.09
candidate_score -> 0.80 * upcode_pressure + 0.10 * upcode_gain
    - 1.10 * audit_penalty - 0.95 * ethics_pressure
    - 0.25 * coding_gap - 100.0 * I(coding_gap > 0.20)

# Iteration 273: delta relative to iteration 213.
bonus_pool -> clip(1.35 + 0.012 * max(0.0, profit_last)
    - 0.60 * I((profit_last < 0.0) or (rbar > 0.08)), 0.75, 3.75)
wrej -> 1.15
triage_accept: urgency 0.16 -> 0.22; wait 0.04 -> 0.05;
               add 0.03 * complexity
candidate_score -> 0.70 * upcode_pressure + 0.12 * upcode_gain
    - 1.15 * audit_penalty - 1.00 * ethics_pressure
    - 0.35 * coding_gap - 100.0 * I(coding_gap > 0.20)
\end{lstlisting}

\section{Implementation Details and DSL Guardrails}\label{sec:app-dsl}
L1 and L2 use the native hospital administrator and provider rules of \S\ref{sec:env}; no \alphaevolve policy is injected into those layers. The main L2 ablation sweeps audit probability, bonus pool, bonus sharpness, total capacity, the health-to-cost hospital \kpi weight ratio, and flexible capacity; the static-flex diagnostic of \S\ref{sec:app-l2} isolates the flexible pool under fixed routing and specialization. L3 selects candidates on validation seeds and reports held-out test results on disjoint seed splits. Unlike L1/L2, L3 may edit selected doctor-side expression constants, but only through the typed fields in Listing~\ref{lst:dsl-box}; host-side dynamics, audit application, and metric computation remain fixed. The mixed-objective scalarizer is the safety-penalized fitness of \S\ref{sec:mac} with $\lambda_{\mathrm{unsafe}}=0.06$ and $\lambda_{\mathrm{var}}=0.25$, with explicit per-step penalties for high-urgency rejection, insolvency, unsafe up-coding, excessive waiting, and high-complexity deferral.

\paragraph{DSL guardrails.}
The released L3 prompt files use a small replayable interface. Listing~\ref{lst:dsl-box} summarizes the shared guardrails. The full artifacts include the system prompt, the full-rewrite template, and the diff template; the \texttt{medisim\_dsl\_v1} tag identifies the DSL schema and is not itself a simulator mechanism. Candidates can edit only fixed policy expression fields and can read only exposed DSL features plus safe scalar helper calls. Simulator dynamics, patient generation, metric computation, host-side clipping, and feasibility projection are held fixed across candidates, which is what closes the audit-evasion gap discussed under desideratum (ii) of \S\ref{sec:mac}. For coding, the host computes \texttt{audit\_penalty} from $p_{\mathrm{audit}}(\Delta\mathrm{CMI})$ and the fixed penalty schedule before the candidate expression is evaluated; \texttt{candidate\_score} may change how strongly the coder responds to that feature, but cannot change the audit probability, penalty application, or reported metrics.

\begin{lstlisting}[caption={L3 DSL guardrail summary.},label={lst:dsl-box},basicstyle=\ttfamily\tiny]
One executable/testable assignment-only medisim_dsl_v1
module, or exact SEARCH/REPLACE blocks.
Top-level: kind, version, admin_policy, doctor_policy.
Admin expr: alpha, beta, total_capacity, bonus_pool,
flex_pool, xi, wH, wW, wrej, wC, kappa.
Doctor expr: effort, triage_accept, triage_defer,
request_bed, candidate_score.
Calls: I, clip, min, max, abs, round, soft_gt, soft_lt.
No new state/schema fields/mechanisms/hidden calls.
Host clips/projects outputs; dynamics and metrics fixed.
\end{lstlisting}

\section{Terminology and Arrival-Process Note}\label{sec:app-terms-arrivals}
\paragraph{Glossary.}
The main text keeps the terminology short; this appendix gives the operational meanings used in the simulator.
\begin{description}[leftmargin=1.2cm,style=nextline]
  \item[Hospital \drg.]
  Diagnosis-related group: a billable case category used in prospective hospital payment. In \medisim, each patient has a true clinical group and a reported group used for settlement.
  \item[Hospital \drg-style arrivals.]
  Hospital patient arrivals with clinical type, urgency, tolerance, and reimbursement-relevant case weight. The phrase distinguishes the stream from identical jobs in a generic queue.
  \item[Hospital \cmi.]
  Case-mix index. We use normalized hospital \cmi as the patient's clinical complexity and as the payment-relevant weight that coding can distort.
  \item[Hospital administrator.]
  The leader in the Stackelberg formulation. This actor sets hospital-level levers such as incentives, audit intensity, capacity, bonus pools, \kpi weights, routing, and steering.
  \item[Hospital Stackelberg game.]
  A leader-follower model in which the hospital administrator commits to a mechanism and hospital providers respond strategically to it.
  \item[Hospital coder and hospital coding policy.]
  A hospital coder maps clinical evidence to a reported billable group. The hospital coding policy is the fixed or searched rule that scores candidate reported groups before the configured coding choice rule.
  \item[Hospital \kpi.]
  A measured hospital performance score assembled from health, waiting, rejection, and cost. It affects bonuses and can diverge from true clinical value.
  \item[Hospital provider-response channels.]
  Up-coding changes the reported billing group; selection changes which patients are accepted; delay keeps patients waiting when serving them is unattractive; effort changes treatment intensity; triage is the local accept/defer/reject and resource-request rule.
  \item[Hospital coding wedge and hospital measurement wedge.]
  The coding wedge is the gap between true clinical complexity and the reported billing group. The measurement wedge is the gap between true clinical value and the measured \kpi used for bonuses or steering.
  \item[Hospital \kpi targeting (Goodhart-style proxy gaming).]
  Provider behavior that improves a measured hospital \kpi while moving away from the hospital's true clinical or access objective.
  \item[Gold-plating, skimping, and cream-skimming.]
  Gold-plating is excessive treatment intensity, skimping is under-provision of care effort or resources, and cream-skimming is accepting easier or more profitable patients disproportionately.
  \item[Hospital Identify--Produce--Settle (IPS).]
  The simulator order: identify patients and billing groups, produce care under capacity constraints, then settle reimbursement, \kpi scores, and bonuses.
  \item[Hospital policy DSL.]
  The restricted domain-specific language used to write candidate hospital policies over approved administrative levers and provider-response expressions.
  \item[Flexible hospital capacity pool and \kpi steering.]
  The flexible capacity pool is capacity that can be reallocated across hospital care teams. \kpi steering assigns that capacity or routing priority using measured hospital performance scores.
\end{description}

\paragraph{Arrival-process scope note.}
The experiments use the homogeneous Poisson law in Eq.~\eqref{eq:arrival-poisson} because it is transparent and reproducible. The simulator does not rely on Poisson arrivals for the channel definitions. Let $A_t$ denote the full arrival batch at period $t$, including patient count and attributes, and let $h_{t-1}$ be the realized pre-period history. For any fixed arrival kernel $\mathcal K_t(dA_t\mid h_{t-1})$ sampled before provider actions, the one-step rollout law under policy $\pi$ factors as
\begin{equation}
\begin{aligned}
&\Pr_\pi(dA_t,dY_t,dX_{t+1}\mid h_{t-1})\\
&\quad=\mathcal K_t(dA_t\mid h_{t-1})\,
\Pr_\pi(dY_t,dX_{t+1}\mid h_{t-1},A_t),
\end{aligned}
\end{equation}
where $Y_t$ collects the identify, routing, triage, treatment, and settlement decisions. The second factor is the strategic hospital-response part studied in the paper. Replacing the Poisson kernel with a nonhomogeneous Poisson process, an over-dispersed count model, correlated service-line arrivals, seasonal mixtures, or an empirical bootstrap stream changes the distribution of states the policy sees, but it leaves the IPS order and the definitions of coding, selection, delay, effort, and \kpi targeting intact, provided the same arrival kernel is used for the policies being compared.

This is the scope condition. If patient demand responds directly to the hospital policy or to reputation generated by earlier policies, the arrival kernel is no longer fixed. That feedback can be modeled as an additional demand-response channel, but it is outside the present L1/L2/L3 experiments.

\end{document}